\pdfoutput=1
\documentclass{article}

\usepackage{microtype}
\usepackage{graphicx}
\usepackage{subfigure}
\usepackage{booktabs} 

\usepackage{hyperref}

\usepackage[accepted]{icml2024}

\usepackage{amsmath}
\usepackage{amssymb}
\usepackage{mathtools}
\usepackage{amsthm}

\usepackage{hyperref}
\usepackage{url}
\usepackage{booktabs}
\usepackage{multicol}
\usepackage{multirow}
\usepackage{graphicx}
\usepackage{subcaption}
\usepackage{colortbl}
\usepackage{xcolor}
\usepackage[most]{tcolorbox}
\usepackage{tabulary}

\usepackage{subfigure}

\usepackage[capitalize,noabbrev]{cleveref}

\theoremstyle{plain}
\newtheorem{theorem}{Theorem}[section]
\newtheorem{proposition}[theorem]{Proposition}
\newtheorem{lemma}[theorem]{Lemma}
\newtheorem{corollary}[theorem]{Corollary}
\theoremstyle{definition}

\theoremstyle{remark}
\newtheorem{remark}[theorem]{Remark}

\usepackage[textsize=tiny]{todonotes}

\icmltitlerunning{Improving Diffusion Models for Inverse Problems Using Optimal Posterior Covariance}

\begin{document}

\twocolumn[
\icmltitle{Improving Diffusion Models for Inverse Problems \\Using Optimal Posterior Covariance}



\icmlsetsymbol{equal}{*}

\begin{icmlauthorlist}
\icmlauthor{Xinyu Peng}{yyy}
\icmlauthor{Ziyang Zheng}{yyy}
\icmlauthor{Wenrui Dai}{yyy}
\icmlauthor{Nuoqian Xiao}{yyy}
\icmlauthor{Chenglin Li}{yyy}
\icmlauthor{Junni Zou}{yyy}
\icmlauthor{Hongkai Xiong}{yyy}
\end{icmlauthorlist}

\icmlaffiliation{yyy}{School of Electronic Information and Electrical Engineering, Shanghai Jiao Tong University, Shanghai, China}

\icmlcorrespondingauthor{Ziyang~Zheng}{zhengziyang@sjtu.edu.cn}
\icmlcorrespondingauthor{Wenrui~Dai}{daiwenrui@sjtu.edu.cn}
\icmlcorrespondingauthor{Junni~Zou}{zoujunni@sjtu.edu.cn}

\icmlkeywords{Machine Learning, ICML}

\vskip 0.3in
]



\printAffiliationsAndNotice{}  

\begin{abstract}
Recent diffusion models provide a promising zero-shot solution to noisy linear inverse problems without retraining for specific inverse problems. In this paper, we reveal that recent methods can be uniformly interpreted as employing a Gaussian approximation with hand-crafted isotropic covariance for the intractable denoising posterior to approximate the conditional posterior mean. Inspired by this finding, we propose to improve recent methods by using more principled covariance determined by maximum likelihood estimation. To achieve posterior covariance optimization without retraining, we provide general plug-and-play solutions based on two approaches specifically designed for leveraging pre-trained models with and without reverse covariance. We further propose a scalable method for learning posterior covariance prediction based on representation with orthonormal basis. Experimental results demonstrate that the proposed methods significantly enhance reconstruction performance without requiring hyperparameter tuning.
\end{abstract}

\section{Introduction}
Noisy linear inverse problems are widely studied for a variety of image processing tasks, including denoising, inpainting, deblurring, super-resolution, among others. The noisy linear inverse problems are formulated to accommodate the widely adopted degradation model, where images are measured with a linear projection under noise corruption.

Recently, diffusion models have been emerging as promising methods for solving inverse problems. According to the training strategies, these diffusion-based solvers can be categorized into two groups: 1) \textit{supervised methods} that aim to learn a conditional diffusion model using datasets consisting of pairs of degraded and clean images~\cite{saharia2022image, whang2022deblurring, luo2023refusion, chan2023sud}, and 2) \textit{zero-shot methods} that leverage pre-trained unconditional diffusion models for conditional sampling in various scenarios of inverse problems without the requirement of retraining. In this paper, we focus on the zero-shot methods that can accommodate various tasks without retraining. To ensure 
the consistency of optimizing fidelity under conditional sampling, existing zero-shot methods adopt projection onto the measurement subspace~\cite{choi2021ilvr, lugmayr2022repaint, song2022solving, wang2023zeroshot, zhu2023denoising}, leverage the similar idea as classifier guidance~\cite{song2021scorebased, dhariwal2021diffusion} to modify the sampling process with the likelihood score~\cite{song2023pseudoinverseguided, chung2023diffusion}, or resort to variational inference~\cite{RN320, mardani2023variational}.

Diffusion models initially establish a forward process that introduces noise to the original data $\mathbf{x}_0$ to generate noisy data $\mathbf{x}_t$ at time $t$. Subsequently, a reverse process generates $\mathbf{x}_0$ following the distribution of the original data. The key to realizing the reverse process lies in the posterior mean $\mathbb{E}[\mathbf{x}_0|\mathbf{x}_t]$, which represents the optimal estimate of $\mathbf{x}_0$ given $\mathbf{x}_t$ in the sense of minimum mean square error (MMSE). In inverse problems, achieving conditional sampling that ensures samples consistent with the measurement $\mathbf{y}$ requires considering the conditional posterior mean $\mathbb{E}[\mathbf{x}_0|\mathbf{x}_t,\mathbf{y}]$, which plays a similar role to $\mathbb{E}[\mathbf{x}_0|\mathbf{x}_t]$ in unconditional sampling. 

However, differing from the methods originated from approximating the conditional score $\nabla_{\mathbf{x}_t} \log p_t(\mathbf{x}_t|\mathbf{y})$ that can be regarded as explicitly approximating $\mathbb{E}[\mathbf{x}_0|\mathbf{x}_t,\mathbf{y}]$~\cite{song2023pseudoinverseguided, chung2023diffusion}, the underlying motivations of recent methods can differ significantly from the objective of approximating $\mathbb{E}[\mathbf{x}_0|\mathbf{x}_t,\mathbf{y}]$. These methods achieve conditional sampling by refining the null space during the reverse diffusion process~\cite{wang2023zeroshot}, interpreting denoising diffusion models as plug-and-play image priors~\cite{zhu2023denoising}, or optimizing the data consistency in the tangent space~\cite{chung2023fast}. There remain two important problems unresolved, \emph{i.e.}, \textit{i) unified interpretation in the sense of approximating the conditional posterior mean}, and \textit{ii) better approximation for the conditional posterior mean}.

In this paper, we address these two problems. We reveal that recent zero-shot methods~\cite{song2023pseudoinverseguided,chung2023diffusion, wang2023zeroshot, zhu2023denoising} can be uniformly interpreted as employing an isotropic Gaussian approximation for the intractable denoising posterior $p_t(\mathbf{x}_0|\mathbf{x}_t)$ to approximate $\mathbb{E}[\mathbf{x}_0|\mathbf{x}_t,\mathbf{y}]$. This perspective allows us to not only reveal a unified interpretation for these methods but also extend the design space of diffusion-based solvers. Specifically, we propose a generalized method to improve recent methods by optimizing posterior covariance based on maximum likelihood estimation (MLE). Numerically, we achieve plug-and-play posterior covariance optimization using pre-trained unconditional diffusion models by converting (available) reverse covariance or via Monte Carlo estimation without reverse covariance. To overcome the quadratic complexity of covariance prediction, we further propose a scalable method for learning posterior covariance prediction by leveraging widely-used orthonormal basis for image processing~(\emph{e.g.}, DCT and DWT basis). Experimental results demonstrate that the proposed method significantly outperforms existing methods in a wide range of tasks, including inpainting, deblurring, and super-resolution, and eliminates the need for hyperparameter tuning.

\section{Background}
\subsection{Bayesian Framework for Solving Inverse Problems}
We consider the linear inverse problems for infering $\mathbf{x}_0 \in \mathbb{R}^d$ from noisy measurements $\mathbf{y}\in \mathbb{R}^m$:
\begin{equation}
\label{eq:forward-model}
    \mathbf{y} = \mathbf{A}\mathbf{x}_0 + \mathbf{n},
\end{equation}
where $\mathbf{A}\in \mathbb{R}^{m\times d}$ is known and $\mathbf{n}\sim \mathcal{N}(\mathbf{0}, \sigma^2 \mathbf{I})$ is an i.i.d. additive Gaussian noise with a known standard deviation of $\sigma$. This gives a likelihood function $p(\mathbf{y}|\mathbf{x}_0)=\mathcal{N}(\mathbf{y}|\mathbf{A} \mathbf{x}_0, \sigma^2 \mathbf{I})$. Under the Bayesian framework, we assume $\mathbf{x}_0$ obeys an unknown prior distribution $p(\mathbf{x}_0)$, and the inverse problems are solved by formulating the posterior distribution over $\mathbf{x}_0$ given $\mathbf{y}$ by Bayes' theorem: $p(\mathbf{x}_0|\mathbf{y}) = p(\mathbf{x}_0) p(\mathbf{y}|\mathbf{x}_0) / \int p(\mathbf{x}_0) p(\mathbf{y}|\mathbf{x}_0) \mathrm{d}\mathbf{x}_0$.

\subsection{Diffusion Models and Conditioning}
\label{sec:intro dpm}
We are interested in using diffusion models to model the complex posterior distribution $p(\mathbf{x}_0|\mathbf{y})$ for inverse problems. Let us define a family of Gaussian perturbation kernels $p_t(\mathbf{x}_t|\mathbf{x}_0)$ of $\mathbf{x}_0\sim p(\mathbf{x}_0)$ by injecting \emph{i.i.d.} Gaussian noise of standard deviation $\sigma_t$ to $\mathbf{x}_0$ and scaling by the factor of $s_t$, \emph{i.e.}, $p_t(\mathbf{x}_t|\mathbf{x}_0) = \mathcal{N}(\mathbf{x}_t|s_t \mathbf{x}_0, s_t^2 \sigma_t^2 \mathbf{I})$, where $\sigma_t$ is monotonically increasing with respect to time $t=0,1,\cdots, T$. We start from $\sigma_0 = 0$ and reach a value of $\sigma_T$ that is much larger than the standard deviation of $p(\mathbf{x}_0)$, to ensure that samples from $\mathbf{x}_T \sim p(\mathbf{x}_T)$ are indistinguishable to samples from $\mathcal{N}(\mathbf{0}, s_T^2 \sigma_T^2 \mathbf{I})$. Since $\mathbf{x}_t$ is independent of $\mathbf{y}$ once $\mathbf{x}_0$ is known, we characterize the joint distribution between $\mathbf{x}_0, \mathbf{y}$, and $\mathbf{x}_t$ as $p_t(\mathbf{x}_0,\mathbf{y},\mathbf{x}_t) = p(\mathbf{x}_0)p(\mathbf{y}|\mathbf{x}_0)p_t(\mathbf{x}_t|\mathbf{x}_0)$,
which can also be represented by a probabilistic graphical model $\mathbf{y} \leftarrow \mathbf{x}_0 \rightarrow \mathbf{x}_t$. There exist multiple formulations of diffusion models in the literature~\cite{song2019generative, ho2020denoising, song2021denoising, kingma2021variational, song2021scorebased, karras2022elucidating}. Here, we use the ordinary differential equation (ODE) formulation and select $s_t=1, \sigma_t = t$ as suggested in~\cite{karras2022elucidating} for simplicity. Let us consider the following ODE:
\begin{equation}
\label{eq:uncond samp}   
    \mathrm{d}\mathbf{x}_t = \frac{\mathbf{x}_t - \mathbb{E}[\mathbf{x}_0|\mathbf{x}_t]}{t} \mathrm{d}t, \ \ \mathbf{x}_T \sim p_T(\mathbf{x}_T).
\end{equation}
In Eq.~\eqref{eq:uncond samp}, the only source of randomness is the initial sample $\mathbf{x}_T \sim p_T(\mathbf{x}_T)$. The ODE possesses an important property that $\mathbf{x}_t$ generated by the ODE maintains the exact same marginals to $\mathbf{x}_t$ obtained by injecting Gaussian noise to $\mathbf{x}_0$, \emph{i.e.}, $p_t(\mathbf{x}_t)$. Generally, diffusion models approximate $\mathbb{E}[\mathbf{x}_0|\mathbf{x}_t]$ with a time-dependent denoiser $D_t(\mathbf{x}_t)$~(referred to as the unconditional diffusion model throughout the paper) that is trained by minimizing the $L_2$ loss for all $t\in [0, T]$:
\begin{equation}  
\min_{D_t}\mathbb{E}_{p_t(\mathbf{x}_0, \mathbf{x}_t)}\left[\lVert \mathbf{x}_0 - D_t(\mathbf{x}_t)\rVert_2^2\right].
\end{equation}
With sufficient data and model capacity, the optimal $D_t(\mathbf{x}_t)$ is the MMSE estimator of $\mathbf{x}_0$ given $\mathbf{x}_t$ and equals to $\mathbb{E}[\mathbf{x}_0|\mathbf{x}_t]$. Thus, samples from $p(\mathbf{x}_0)$ can be obtained by first sampling $\mathbf{x}_T$ from $\mathcal{N}(\mathbf{0}, s_T^2 \sigma_T^2 \mathbf{I})$ and then simulating Eq.~\eqref{eq:uncond samp} from $t = T$ to $t = 0$ using black box ODE solver that replaces $\mathbb{E}[\mathbf{x}_0|\mathbf{x}_t]$ with a well-trained $D_t(\mathbf{x}_t)$.


To solve inverse problems, we are interested in $p(\mathbf{x}_0|\mathbf{y})$, which requires constructing an ODE whose marginals are $p_t(\mathbf{x}_t|\mathbf{y})$. The desired ODE is
\begin{equation}
\label{eq: cond sample}
    \mathrm{d}\mathbf{x}_t = \frac{\mathbf{x}_t - \mathbb{E}[\mathbf{x}_0|\mathbf{x}_t, \mathbf{y}]}{t} \mathrm{d}t, \ \ \mathbf{x}_T \sim p_T(\mathbf{x}_T|\mathbf{y}).
\end{equation}
For sufficiently large $\sigma_T$, samples $\mathbf{x}_T \sim p(\mathbf{x}_T|\mathbf{y})$ are indistinguishable to samples from $\mathcal{N}(\mathbf{0}, s_T^2 \sigma_T^2 \mathbf{I})$~(see Appendix D.2 in \cite{dhariwal2021diffusion}). The unconditioned sampling procedure in Eq.~\eqref{eq:uncond samp} can be used to achieve conditioned sampling in Eq.~\eqref{eq: cond sample}, by substituting $\mathbb{E}[\mathbf{x}_0|\mathbf{x}_t]$ with the conditional posterior mean  $\mathbb{E}[\mathbf{x}_0|\mathbf{x}_t, \mathbf{y}]$\footnote{Note that this conclusion can be extended to DDPM, DDIM, and SDE-formalized diffusion models.}.

\section{Unified Interpretation of Diffusion-based Solvers to Inverse Problems}\label{sec:unif}
To simulate Eq.~\eqref{eq: cond sample} for solving inverse problems, a standard approach is training a conditional diffusion model for estimating $\mathbb{E}[\mathbf{x}_0|\mathbf{x}_t, \mathbf{y}]$ using supervised learning. However, this approach can be computationally demanding by training separate models for different scenarios. In this paper, we focus on zero-shot methods, which leverage pre-trained unconditional diffusion models for conditional sampling in inverse problems to avoid additional training. We provide a construction of zero-shot methods based on conditional posterior mean approximation as below: \\
\noindent 1) \textbf{Denoising} Obtain an estimation of $\mathbb{E}[\mathbf{x}_0|\mathbf{x}_t]$ through the unconditional diffusion model $D_t(\mathbf{x}_t)$; \\
\noindent 2) \textbf{Modification} Obtain an approximation $\hat{\mathbf{x}}_0^{(t)}$ for the conditional posterior mean $\mathbb{E}[\mathbf{x}_0|\mathbf{x}_t, \mathbf{y}]$ based on $\mathbf{y}$ and $D_t(\mathbf{x}_t)$; \\
\noindent 3) \textbf{Substitution} Substitute $\hat{\mathbf{x}}_0^{(t)}$ for $\mathbb{E}[\mathbf{x}_0|\mathbf{x}_t]$ in the unconditional sampling process Eq.~\eqref{eq:uncond samp}.

In this section, we unify existing methods with the interpretation of approximating $\mathbb{E}[\mathbf{x}_0|\mathbf{x}_t, \mathbf{y}]$ with isotropic Gaussian approximations $\mathcal{N}(\mathbf{x}_0|D_t(\mathbf{x}_t), r_t^2 \mathbf{I})$ to the intractable denoising posterior $p_t(\mathbf{x}_0|\mathbf{x}_t)$ with different $r_t$, as summarized in Table~\ref{tab:compare}. We first review the Type I guidance methods that explicitly approximate $\mathbb{E}[\mathbf{x}_0|\mathbf{x}_t,\mathbf{y}]$, including DPS~\cite{chung2023diffusion} and $\Pi\text{GDM}$~\cite{song2023pseudoinverseguided}. Subsequently, we reveal that the proximal-based methods like DDNM~\cite{wang2023zeroshot} and DiffPIR~\cite{zhu2023denoising} can also be interpreted as approximating $\mathbb{E}[\mathbf{x}_0|\mathbf{x}_t,\mathbf{y}]$ and categorize them into the Type II guidance methods. 

\subsection{Type I Guidance: Approximating the Likelihood Score Function}\label{sec:tIguid}
We classify DPS and $\Pi\text{GDM}$ into the category of Type I guidance that originates from approximating the conditional score $\nabla_{\mathbf{x}_t} \log p_t(\mathbf{x}_t|\mathbf{y}) = \nabla_{\mathbf{x}_t} \log p_t(\mathbf{x}_t) + \nabla_{\mathbf{x}_t} \log p_t(\mathbf{y}|\mathbf{x}_t)$. We represent these methods with an equivalent form of conditional posterior mean approximation based on Proposition~\ref{prop:exy}.
\begin{proposition}\label{prop:exy} 
The conditional posterior mean is equal to the posterior mean drifted by scaled likelihood score function. Formally,
\begin{equation}\label{eq.type1-cpm}
\mathbb{E}[\mathbf{x}_0|\mathbf{x}_t,\mathbf{y}] = \mathbb{E}[\mathbf{x}_0|\mathbf{x}_t] + s_t\sigma^2_{t} \nabla_{\mathbf{x}_t}\log p_t(\mathbf{y}|\mathbf{x}_t).
\end{equation}
\begin{proof}
Please refer to Appendix~\ref{app:prop.exy}.
\end{proof}
\end{proposition}
To obtain $\hat{\mathbf{x}}_0^{(t)}$ according to Proposition~\ref{prop:exy}\footnote{In addition to realizing guidance for diffusion models, Proposition~\ref{prop:exy} can be used for refining the estimation of $\mathbf{x}_0$ for learning sampling patterns in MRI~\citep{ravula2023optimizing} and realizing guidance for flow-based generative models to solve inverse problems~\citep{pokle2023training}.}, note that $\mathbb{E}[\mathbf{x}_0|\mathbf{x}_t]$ can be estimated using $D_t(\mathbf{x}_t)$; however, the likelihood score $\nabla_{\mathbf{x}_t}\log p_t(\mathbf{y}|\mathbf{x}_t)$ is computationally intractable. In inverse problems, only the likelihood $p(\mathbf{y}|\mathbf{x}_0)$ at $t=0$ is known, and $p_t(\mathbf{y}|\mathbf{x}_t)$ for any $t>0$ is obtained by an intractable integral over all possible $\mathbf{x}_0$ as
\begin{equation}
p_t(\mathbf{y}|\mathbf{x}_t) = \int p(\mathbf{y}|\mathbf{x}_0)p_{t}(\mathbf{x}_0|\mathbf{x}_t) \mathrm{d}\mathbf{x}_0.
\end{equation}
Thus, one possible way to obtain $\mathbb{E}[\mathbf{x}_0|\mathbf{x}_t, \mathbf{y}]$ is to consider approximating $p_t(\mathbf{y}|\mathbf{x}_t)$. DPS and $\Pi\text{GDM}$ achieve this by uniformly employing isotropic Gaussian approximation for $p_{t}(\mathbf{x}_0|\mathbf{x}_t)$, as elaborated below.

\begin{table}[!t]
\renewcommand{\baselinestretch}{1.0}
\renewcommand{\arraystretch}{1.0}
\centering
\begin{tabular}{@{}l|c|c@{}}
\toprule
\multicolumn{1}{c|}{Methods} & Guidance & $r_t$ \\
\midrule
DPS~\cite{chung2023diffusion} & I & $\text{approach}~0$ \\
$\Pi\text{GDM}$~\cite{song2023pseudoinverseguided} & I & $\sqrt{\sigma_t^2 / (\sigma_t^2 + 1)}$ \\
DDNM~\cite{wang2023zeroshot} & II & any fixed value \\
DiffPIR~\cite{zhu2023denoising} & II & $\sigma_t / \sqrt{\lambda}$ \\
\bottomrule
\end{tabular}
\caption{\textbf{Unified interpretation of diffusion-based solvers to inverse problems.} Recent methods can be regarded as making isotropic Gaussian approximations to the denoising posterior. $p_t(\mathbf{x}_0|\mathbf{x}_t)$}\label{tab:compare}
\end{table}

\textbf{DPS~\cite{chung2023diffusion}} DPS can be viewed as approximating $p_{t}(\mathbf{x}_0|\mathbf{x}_t)$ using a delta distribution $\delta(\mathbf{x}_0-D_t(\mathbf{x}_t))$ centered at the posterior mean estimate $D_t(\mathbf{x}_t)$, which can be regarded as the limit of the Gaussian $\mathcal{N}(\mathbf{x}_0|D_t(\mathbf{x}_t), r_t^2 \mathbf{I})$ when the variance $r_t^2$ approaches zero. The likelihood $p_t(\mathbf{y}|\mathbf{x}_t)$ is approximated by
\begin{align}
p_t(\mathbf{y}|\mathbf{x}_t) &\approx \int p(\mathbf{y}|\mathbf{x}_0)\delta(\mathbf{x}_0-D_t(\mathbf{x}_t)) \mathrm{d}\mathbf{x}_0 \nonumber\\
&= p(\mathbf{y}|\mathbf{x}_0=D_t(\mathbf{x}_t)). \label{eq:dps}
\end{align}
However, directly using Eq.~\eqref{eq:dps} does not perform well in practice, and \citet{chung2023diffusion} empirically adjusts the strength of guidance by approximating the likelihood score $\nabla_{\mathbf{x}_t}\log p_t(\mathbf{y}|\mathbf{x}_t)$ with $-\zeta_t \nabla_{\mathbf{x}_t} \lVert \mathbf{y} - \mathbf{A} D_t(\mathbf{x}_t)\rVert_2^2$, where $\zeta_t = \zeta / \lVert \mathbf{y} - \mathbf{A} D_t(\mathbf{x}_t)\rVert_2$ with a hyper-parameter $\zeta$. 

\textbf{$\Pi\text{GDM}$~\cite{song2023pseudoinverseguided}} The delta distribution used in DPS is a very rough approximation to $p_t(\mathbf{x}_0|\mathbf{x}_t)$ as it completely ignores the uncertainty of $\mathbf{x}_0$ given $\mathbf{x}_t$. 
As $t$ increases, the uncertainty in $p_t(\mathbf{x}_0|\mathbf{x}_t)$ becomes larger and is closed to the original data distribution $p(\mathbf{x}_0)$. Thus, it is more reasonable to choose a positive $r_t$.
In $\Pi\text{GDM}$, $r_t$ is heuristically selected as $\sqrt{\sigma_t^2/(1 + \sigma_t^2)}$ under the assumption that $p(\mathbf{x}_0)$ is the standard normal distribution $\mathcal{N}(\mathbf{0}, \mathbf{I})$. In such case, the likelihood $p_t(\mathbf{y}|\mathbf{x}_t)$ is approximated by
\begin{align}
p_t(\mathbf{y} |\mathbf{x}_t) &\approx\int\mathcal{N}(\mathbf{y}|\mathbf{A} \mathbf{x}_0, \sigma^2 \mathbf{I}) \mathcal{N}(\mathbf{x}_0|D_t(\mathbf{x}_t), r_t^2 \mathbf{I}) \mathrm{d}\mathbf{x}_0\nonumber\\
&= \mathcal{N}(\mathbf{y}|\mathbf{A}D_t(\mathbf{x}_t), \sigma^2 \mathbf{I} + r_t^2\mathbf{A} \mathbf{A}^T),\label{eq:pgdm-likelihood}
\end{align}
where the gradient of the log-likelihood of Eq.~\eqref{tab:complete-pgdm} can be computed via Jacobian-vector product using back-propagation for approximating $\nabla_{\mathbf{x}_t}\log p_t(\mathbf{y}|\mathbf{x}_t)$.


\subsection{Type II Guidance: Approximating the Conditional Posterior Mean Using Proximal Solution}\label{sec:tIIguid}
We classify DiffPIR and DDNM into the category of Type II guidance, which replaces $\mathbb{E}[\mathbf{x}_0|\mathbf{x}_t]$ in the unconditional sampling process using proximal solutions. 
Although Type II guidance methods have been developed from  perspectives that differ from approximating $\mathbb{E}[\mathbf{x}_0|\mathbf{x}_t,\mathbf{y}]$, we show that it can also be regarded as approximating $\mathbb{E}[\mathbf{x}_0|\mathbf{x}_t,\mathbf{y}]$ by employing isotropic Gaussian approximation to the denoising posterior.

\textbf{DiffPIR~\cite{zhu2023denoising}} The core step in DiffPIR is replacing $\mathbb{E}[\mathbf{x}_0|\mathbf{x}_t]$ in the unconditional sampling process with the solution to the following proximal problem:
\begin{equation}\label{eq:exy-diffpir}
\hat{\mathbf{x}}_0^{(t)} = \arg\min_{\mathbf{x}_0} \lVert \mathbf{y} -\mathbf{A}\mathbf{x}_0 \rVert^2 +  \rho_t \lVert \mathbf{x}_0 - D_t(\mathbf{x}_t) \rVert^2_2,
\end{equation}
where $\rho_t = \lambda \sigma^2 / \sigma_t^2$ and $\lambda$ is a hyper-parameter. Our key insight is that Eq.~\eqref{eq:exy-diffpir} for DiffPIR can be interpreted as approximating $\mathbb{E}[\mathbf{x}_0|\mathbf{x}_t, \mathbf{y}]$ with the mean of an approximate distribution $q_t(\mathbf{x}_0|\mathbf{x}_t,\mathbf{y})$ derived from the isotropic Gaussian approximation $q_t(\mathbf{x}_0|\mathbf{x}_t)$ to $p_t(\mathbf{x}_0|\mathbf{x}_t)$. Let us define the isotropic Gaussian approximation $q_t(\mathbf{x}_0|\mathbf{x}_t)=\mathcal{N}(\mathbf{x}_0|D_t(\mathbf{x}_t), r_t^2 \mathbf{I})$. We can obtain that the approximate distribution $q_t(\mathbf{x}_0|\mathbf{x}_t,\mathbf{y})\propto p(\mathbf{y}|\mathbf{x}_0)q_t(\mathbf{x}_0|\mathbf{x}_t)$ for $p_t(\mathbf{x}_0|\mathbf{x}_t,\mathbf{y}) \propto p(\mathbf{y}|\mathbf{x}_0)p_t(\mathbf{x}_0|\mathbf{x}_t)$\footnote{$p_t(\mathbf{y}|\mathbf{x}_0, \mathbf{x}_t)=p(\mathbf{y}|\mathbf{x}_0)$ since we leverage the conditional independence between $\mathbf{y}$ and $\mathbf{x}_t$ given $\mathbf{x}_0$.} is also a Gaussian and its mean $\mathbb{E}_q[\mathbf{x}_0|\mathbf{x}_t,\mathbf{y}]$ can be obtained by solving the optimization problem:
\begin{align}\label{eq:proximal}
&\mathbb{E}_q[\mathbf{x}_0|\mathbf{x}_t,\mathbf{y}] = \arg\max_{\mathbf{x}_0} \log q_t(\mathbf{x}_0|\mathbf{x}_t,\mathbf{y}) \nonumber\\
&\ = \arg\max_{\mathbf{x}_0} \left[\log p(\mathbf{y}|\mathbf{x}_0)+\log q_t(\mathbf{x}_0|\mathbf{x}_t)\right] \nonumber\\
&\ = \arg\min_{\mathbf{x}_0} \left[\lVert\mathbf{y}-\mathbf{A}\mathbf{x}_0 \rVert^2+\frac{\sigma^2}{r_t^2}\lVert\mathbf{x}_0-D_t(\mathbf{x}_t) \rVert^2_2\right].
\end{align}
Note that Eq.~\eqref{eq:exy-diffpir} used in DiffPIR is a special case of Eq.~\eqref{eq:proximal} with $r_t=\sigma_t/\sqrt{\lambda}$. Therefore, DiffPIR can be viewed as using isotropic Gaussian approximation $q_t(\mathbf{x}_0|\mathbf{x}_t)=\mathcal{N}(\mathbf{x}_0|D_t(\mathbf{x}_t),(\sigma_t^2/\lambda)\mathbf{I})$ for $p_t(\mathbf{x}_0|\mathbf{x}_t)$.

\textbf{DDNM~\cite{wang2023zeroshot}} The core step in DDNM resorts to the range-null space decomposition of $D_t(\mathbf{x}_t)$. Specifically, to ensure the measurement consistency $\mathbf{y} = \mathbf{A} \hat{\mathbf{x}}_0^{(t)}$ of the solution $\hat{\mathbf{x}}_0^{(t)}$, DDNM formulates $\hat{\mathbf{x}}_0^{(t)}=\mathbf{A}^{\dagger}\mathbf{y} + (\mathbf{I} - \mathbf{A}^{\dagger}\mathbf{A})D_t(\mathbf{x}_t)$ by replacing the range space component of $D_t(\mathbf{x}_t)$ with $\mathbf{A}^{\dagger}\mathbf{y}$ but keeping the null space component unchanged. We find that, when the measurement noise $\sigma$ vanishes, replacing the range space component $\mathbf{A}^{\dagger}\mathbf{A}D_t(\mathbf{x}_t)$ with $\mathbf{A}^{\dagger}\mathbf{y}$ is equivalent to replacing $D_t(\mathbf{x}_t)$ with $\mathbb{E}_q[\mathbf{x}_0|\mathbf{x}_t,\mathbf{y}]$. Therefore, DDNM is equivalent to approximating $p_t(\mathbf{x}_0|\mathbf{x}_t)$ using an isotropic Gaussian, as formalized in Proposition~\ref{prop:ddnm}. 
\begin{proposition}\label{prop:ddnm}
For any $r_t>0$, $\mathbb{E}_q[\mathbf{x}_0|\mathbf{x}_t,\mathbf{y}]$ approaches $\hat{\mathbf{x}}_0^{(t)}$ used in DDNM as the variance of measurement noise approaches zero. Formally, we have
\begin{equation}\label{eq:ddnm}
\lim_{\sigma\rightarrow 0}\mathbb{E}_q[\mathbf{x}_0|\mathbf{x}_t,\mathbf{y}] = \mathbf{A}^{\dagger}\mathbf{y} + (\mathbf{I} - \mathbf{A}^{\dagger}\mathbf{A})D_t(\mathbf{x}_t).
\end{equation}
\begin{proof}
Please refer to~Appendix~\ref{app:proof.ddnm}.
\end{proof}
\end{proposition}

\subsection{Solving Inverse Problems with Optimal Posterior Covariance}
In Sections~\ref{sec:tIguid} and~\ref{sec:tIIguid}, we unify existing zero-shot diffusion-based solvers with the interpretation of isotropic Gaussian approximations to the intractable denoising posterior $p_t(\mathbf{x}_0|\mathbf{x}_t)$. This unified interpretation motivates us to further optimize the posterior covariance in the Gaussian approximation for enhancing existing diffusion-based solvers.
Specifically, we consider to approximate $p_t(\mathbf{x}_0|\mathbf{x}_t)$ using variational Gaussian $q_t(\mathbf{x}_0|\mathbf{x}_t)=\mathcal{N}(\mathbf{x}_0|D_t(\mathbf{x}_t),\Sigma_t(\mathbf{x}_t))$ with learnable covariance $\Sigma_t(\mathbf{x}_t)$. In the pre-training stage, $q_t(\mathbf{x}_0|\mathbf{x}_t)$ is learned by minimizing the weighted integral of expected forward KL divergence between $p_t(\mathbf{x}_0|\mathbf{x}_t)$ and $q_t(\mathbf{x}_0|\mathbf{x}_t)$ as
\begin{equation}\label{eq:klobj}
\min_{q} \int \omega_t \mathbb{E}_{p_t(\mathbf{x}_t)}[ D_{KL}(p_t(\mathbf{x}_0|\mathbf{x}_t)\|q_t(\mathbf{x}_0|\mathbf{x}_t))] \mathrm{d}t. 
\end{equation}
Eq.~\eqref{eq:klobj} can be achieved in a tractable way by maximizing the log-likelihood of $q_t(\mathbf{x}_0|\mathbf{x}_t)$. When $\Sigma_t(\mathbf{x}_t)$ is full rank, the optimal $D_t(\mathbf{x}_t)$ is exactly the MMSE estimator $\mathbb{E}[\mathbf{x}_0|\mathbf{x}_t]$. Once $q_t(\mathbf{x}_0|\mathbf{x}_t)$ is learned, we can leverage $D_t$ and $\Sigma_t$ to solve inverse problems and accommodate existing methods by plugging $D_t(\mathbf{x}_t)$ in~Eq.~\eqref{eq:uncond samp} to achieve unconditional sampling. The solutions to Type I and Type II guidance are summarized as below.

\textbf{Type I guidance.} The likelihood is approximated in a similar way to Eq.~\eqref{eq:pgdm-likelihood} using $D_t$ and $\Sigma_t$:
\begin{equation}
\label{eq:impt1}
p_t(\mathbf{y}|\mathbf{x}_t) \approx \mathcal{N}(\mathbf{y}|\mathbf{A}D_t(\mathbf{x}_t), \sigma^2 \mathbf{I} + \mathbf{A} \Sigma_t(\mathbf{x}_t) \mathbf{A}^T).
\end{equation}

\textbf{Type II guidance.} $\hat{\mathbf{x}}_0^{(t)}$ is solved from the \textit{auto-weighted} proximal problem associated with $D_t$ and $\Sigma_t$:
\begin{equation}\label{eq:impt2}
\hat{\mathbf{x}}_0^{(t)} = \arg\min_{\mathbf{x}_0} \lVert \mathbf{y} -\mathbf{A}\mathbf{x}_0 \rVert^2 + \sigma^2 \lVert \mathbf{x}_0 - D_t(\mathbf{x}_t) \rVert^2_{\Sigma_t^{-1}},
\end{equation}
where $\lVert \mathbf{x} \rVert_{\Lambda}^2=\mathbf{x}^T \Lambda \mathbf{x}$ with positive definite matrix $\Lambda$. 

To avoid the inversion of high-dimensional matrices for realizing guidances in practice, we develop efficient closed-form solutions under the isotropic posterior covariance $\Sigma_t(\mathbf{x}_t) = r_t^2(\mathbf{x}_t) \mathbf{I}$, as well as numerical solutions based on the conjugate gradient method (CG) for more general covariance. Please refer to Appendix~\ref{sec:numerical} for details.

\section{Posterior Covariance Optimization}

In this section, we discuss practical methods for posterior covariance optimization. We propose two plug-and-play methods which can be directly applied to recent methods for two common cases: 1) reverse covariance prediction is available from the given unconditional diffusion model~(Section~\ref{sec:convert}), and 2) reverse covariance prediction is not available~(Section~\ref{sec:analytic}). Furthermore, to model the ubiquitous pixel-correlations in natural images, we propose to learn the variance in the transform space (Section~\ref{sec:ot}), which addresses the quadratic complexity in covariance prediction.

\subsection{Converting Optimal Reverse Variances}\label{sec:convert}
Recent pre-trained diffusion models often predict the optimal \textit{reverse variances\footnote{Variances are the diagonal elements of the covariance matrix.}}~\cite{nichol2021improved} for improving performance when using ancestral sampling of denoising diffusion probabilistic models~(DDPM,~\cite{ho2020denoising, sohl2015deep}). In this section, we reveal that the reverse variance prediction can be leveraged for posterior variance prediction. To achieve this, We first develop the fixed-point solutions to Eq.~\eqref{eq:klobj} in Proposition~\ref{prop:opt-klobj} and then establish the relation between the fixed-point solutions to Eq.~\eqref{eq:klobj} and DDPM in Eq.~\eqref{eq:vae-elbo0} in Theorem~\ref{thm:relation of pos and rev var}. 
\begin{proposition}[Fixed-point solutions of variational Gaussian posterior]
\label{prop:opt-klobj}
The optimal mean $D_t^*(\mathbf{x}_t)$ and the optimal diagonal posterior covariance $\Sigma_t^*(\mathbf{x}_t)=\mathrm{diag}[\mathbf{r}_t^{*2}(\mathbf{x}_t)]$ to Eq.~\eqref{eq:klobj} are obtained by
\begin{align}
    &D_t^*(\mathbf{x}_t) = \mathbb{E}[\mathbf{x}_0|\mathbf{x}_t], \\
    &\mathbf{r}_t^{*2}(\mathbf{x}_t) = \mathbb{E}_{p_t(\mathbf{x}_0|\mathbf{x}_t)}[(\mathbf{x}_0-\mathbb{E}[\mathbf{x}_0|\mathbf{x}_t])^2], \label{eq:opt-pos-var} 
\end{align}
\begin{proof}
Please refer to Appendix~\ref{app:proof.opt-klobj}.
\end{proof}
\end{proposition}
where $(\cdot)^2$ denotes element-wise square. Subsequently, we consider the fixed-point solutions of DDPM. Unlike continuous ODE formulation introduced in Section~\ref{sec:intro dpm}, diffusion models pre-trained under the DDPM framework are latent variable models defined by $p(\mathbf{x}_0)=\int p(\mathbf{x}_{0:T}) \mathrm{d}\mathbf{x}_{1:T}$. The joint distribution $p(\mathbf{x}_{0:T})$ is referred to as the \textit{reverse process} defined as a Markov chain of learnable Gaussian transitions starting at $p(\mathbf{x}_T)=\mathcal{N}(\mathbf{0},\mathbf{I})$, which are characterized by the mean $\mathbf{m}_t$ and covariance $\mathbf{C}_t$:
\begin{align}
&p(\mathbf{x}_{0:T}) = p(\mathbf{x}_T)\prod_{t=1}^T p(\mathbf{x}_{t-1}|\mathbf{x}_t), \\
&p(\mathbf{x}_{t-1}|\mathbf{x}_t) = \mathcal{N}(\mathbf{x}_{t-1}|\mathbf{m}_t(\mathbf{x}_t), \mathbf{C}_t(\mathbf{x}_t)).
\end{align}

DDPM defines a forward process $q(\mathbf{x}_{1:T}|\mathbf{x}_0)$ by gradually injecting noise to the data. Please refer to Appendix~\ref{sec:pf relation of pos and rev var} for the detailed definition of $q(\mathbf{x}_{1:T}|\mathbf{x}_0)$. We fit $p(\mathbf{x}_0)$ to the original data distribution $q(\mathbf{x}_0)$ by minimizing the KL divergence between the forward and reverse processes:
\begin{equation}\label{eq:vae-elbo0} 
    \min_p D_{KL}(q(\mathbf{x}_{0:T}) || p(\mathbf{x}_{0:T})), 
\end{equation}
where $q(\mathbf{x}_{0:T}) = q(\mathbf{x}_0) q(\mathbf{x}_{1:T}|\mathbf{x}_0)$. In Theorem~\ref{thm:relation of pos and rev var}, we present the fixed-point solution of DDPM under diagonal covariance $\mathbf{C}_t(\mathbf{x}_t)=\mathrm{diag}[\mathbf{v}_t^2(\mathbf{x}_t)]$ as used in~\cite{nichol2021improved}.   
\begin{theorem}[Fixed-point solutions of DDPM] 
\label{thm:relation of pos and rev var}
Let $\mathbf{C}_t(\mathbf{x}_t)=\mathrm{diag}[\mathbf{v}_t^2(\mathbf{x}_t)]$ be a signal-dependent diagonal covariance for the reverse covariance. When $\Tilde{\mu}_t, \beta_t, \bar{\alpha}_t, \bar{\beta}_t$ are determined by the forward process $q(\mathbf{x}_{1:T}|\mathbf{x}_0)$, the optimal solutions $\mathbf{m}^*_t(\mathbf{x}_t)$ and $\mathbf{v}_t^{*2}(\mathbf{x}_t)$ to Eq.~\eqref{eq:vae-elbo0} are
\begin{align}
&\mathbf{m}^*_t(\mathbf{x}_t) = \Tilde{\mu}_t(\mathbf{x}_t, \mathbb{E}[\mathbf{x}_0|\mathbf{x}_t]), \\
&\mathbf{v}_t^{*2}(\mathbf{x}_t) = \Tilde{\beta}_t + (\frac{\sqrt{\bar{\alpha}_{t-1}}\beta_t}{1-\bar{\alpha}_t})^{2} \cdot \mathbf{r}_t^{*2}(\mathbf{x}_t) \label{eq:opt-revvar}
\end{align}
where $\mathbf{r}_t^{*2}(\mathbf{x}_t)$ is the optimal posterior variances determined by~Eq.~\eqref{eq:opt-pos-var} under $s_t = \sqrt{\bar{\alpha}_t}$ and $\sigma_t = \sqrt{\bar{\beta}_t/\bar{\alpha}_t}$, and $\Tilde{\beta}_t = (\bar{\beta}_{t-1}/\bar{\beta}_t)\beta_t$.
\begin{proof}
Please refer to Appendix~\ref{sec:pf relation of pos and rev var}.    
\end{proof}
\end{theorem}
Theorem~\ref{thm:relation of pos and rev var} implies that, given the reverse variances $\hat{\mathbf{v}}_t^{2}(\mathbf{x}_t)$ predicted by a pre-trained DDPM model at time $t$, the posterior variances are obtained according to Eq.~\eqref{eq:opt-revvar} by
\begin{equation}
\label{eq:convert}
\hat{\mathbf{r}}_t^{2}(\mathbf{x}_t) = (\hat{\mathbf{v}}_t^{2}(\mathbf{x}_t) - \Tilde{\beta}_t)\cdot\left(\frac{\sqrt{\bar{\alpha}_{t-1}}\beta_t}{1-\bar{\alpha}_t}\right)^{-2}.
\end{equation}

In fact, similar fixed-point solutions of DDPM have been proposed in \cite{bao2022analyticdpm} and \cite{pmlr-v162-bao22d} for determining optimal reverse variances. Our approach can be viewed as opposite direction to \cite{bao2022analyticdpm} and \cite{pmlr-v162-bao22d} for applying the fixed-point solutions.

\subsection{Monte Carlo Estimation of Posterior Variances}\label{sec:analytic}
We further develop a method using a model without providing the reverse variances prediction $\hat{\mathbf{v}}_t^2(\mathbf{x}_t)$~(\emph{e.g.}, \cite{ho2020denoising}). In this case, we suppose signal-independent $\Sigma_t(\mathbf{x}_t) = r_t^2 \mathbf{I}$ is an isotropic covariance with a time-dependent variance $r_t^2$. We directly optimize $r_t^2$ by forcing the derivative of Eq.~\eqref{eq:klobj} w.r.t. $r_t^2$ to zero. Thus, we obtain that
\begin{equation}
r_t^{*2} = \frac{1}{d}\mathbb{E}_{p_t(\mathbf{x}_0,\mathbf{x}_t)}\left[\lVert \mathbf{x}_0 - D_t (\mathbf{x}_t)\rVert^2_2\right].
\end{equation}
Note that $r_t^{*2}$ is the the expected reconstruction error of $D_t (\mathbf{x}_t)$, which can be estimated using Monte Carlo samples to calculate the empirical mean of $\lVert \mathbf{x}_0 - D_t (\mathbf{x}_t)\rVert^2_2$ offline. In the experiments, we pre-compute the empirical mean for 1000 discrete time steps using $5\%$ of the dataset, and for any $t$ used in sampling, we use the pre-computed result with the nearest time step to $t$.

\subsection{Modeling Pixel-Correlations With Latent Variances} \label{sec:ot}
We have achieved posterior covariance optimization in a plug-and-play fashion in Sections~\ref{sec:convert} and~\ref{sec:analytic}. 
In this section, we present a scalable way to model pixel correlations using orthonormal transforms, motivated by transform coding~\cite{goyal2001theoretical} for image compression. We assume $\mathbf{x}_0$ can be represented by some orthonormal basis $\mathbf{\Psi}$, such that $\mathbf{x}_0 = \mathbf{\Psi} \theta_0$. According to the property of covariance,
\begin{equation}\label{eq:otvar}
    \mathrm{Cov}[\mathbf{x}_0|\mathbf{x}_t] = \mathbf{\Psi}\mathrm{Cov}[\theta_0|\mathbf{x}_t] \mathbf{\Psi}^T.
\end{equation}
Here, the elements of $\theta_0$ are supposed to be mutually independent. Since conditioning on  $\mathbf{x}_t$ is equivalent to conditioning on a perturbed version of $\theta_0$ with isotropic Gaussian noise, the elements of $\theta_0|\mathbf{x}_t$ are also mutually independent. This motivates the idea that, $\mathrm{Cov}[\theta_0|\mathbf{x}_t]$ could be better approximated by a diagonal matrix than $\mathrm{Cov}[\mathbf{x}_0|\mathbf{x}_t]$, if the elements of $\theta_0$ are more ``closed to'' mutually independent than $\mathbf{x}_0$ using a proper basis. Thereby, we parameterize the covariance of the variational Gaussian posterior $q_t(\mathbf{x}_0|\mathbf{x}_t)$ as 
\begin{equation}
    \Sigma_t(\mathbf{x}_t) = \mathbf{\Psi} \mathrm{diag}[\mathbf{r}_t^2(\mathbf{x}_t)] \mathbf{\Psi}^T.
\end{equation}
Note that the number of parameters to predict is significantly reduced from $\mathcal{O}(d^2)$ to $\mathcal{O}(d)$, especially for image data with $d>10^5$. The training procedure can be simply implemented as learning a diagonal Gaussian in the transform domain (see Appendix~\ref{app:ot} for details). We adopt the discrete wavelet transform (DWT) for $\mathbf{\Psi}$ in our experiments.

\section{Related Work}

\textbf{Posterior covariance optimization.} Concurrent to our work, \citet{boys2023tweedie} leverage second-order Tweedie's formula to optimize posterior covariance for solving inverse problems. 
The optimal $\Sigma_t(\mathbf{x}_t)$ to~\eqref{eq:klobj} is solved using the Jacobian of $\mathbb{E}[\mathbf{x}_0|\mathbf{x}_t]$, \emph{i.e.}, $\Sigma_t^*(\mathbf{x}_t)=\sigma_t^2\nabla_{\mathbf{x}_t}\mathbb{E}[\mathbf{x}_0|\mathbf{x}_t]$\footnote{According to moment matching~\cite{bishop2006pattern}, $\Sigma_t^*(\mathbf{x}_t)$ equals to the covariance of the denoising posterior $\mathrm{Cov}[\mathbf{x}_0|\mathbf{x}_t]$.}. However, this approach requires strong approximation to scale to high-dimensional data like images. In particular, row sum approximation for the Jacobian, \emph{i.e.}, $\nabla_{\mathbf{x}_t}\mathbb{E}[\mathbf{x}_0|\mathbf{x}_t]\approx \mathrm{diag}[\nabla_{\mathbf{x}_t}\mathbf{1}^TD_t(\mathbf{x}_t)]$ is adopted for image experiments. \citet{rout2023beyond} use only the trace of $\nabla_{\mathbf{x}_t}D_t(\mathbf{x}_t)$. This is related to Section~\ref{sec:analytic}, since the trace of $\Sigma_t^*(\mathbf{x}_t)$ equals to $\mathbb{E}_{p_t(\mathbf{x}_0|\mathbf{x}_t)}[\lVert \mathbf{x}_0 - \mathbb{E}[\mathbf{x}_0|\mathbf{x}_t]\rVert^2_2]$. In contrast, our approach does not introduce additional computational cost for covariance prediction at the inference time.

\textbf{High-order denoising score matching.} The approach presented in Section~\ref{sec:ot} can be viewed as a variant of high-order denoising score matching~\cite{meng2021estimating, lu2022maximum} whose goal is to directly predict the high-order moments of the denoising posterior without exploiting high-order Tweedie's formulas. This is more practically appealing since small error in the first-order loss does not guarantee small estimation error in high-order scores and the Tweedie's approaches could suffer from large estimation error in practice~\cite{meng2021estimating}.  However, these approaches~\cite{meng2021estimating, lu2022maximum} can only be employed on small-scale datasets. To our best knowledge, our approach is the first scalable method applicable for large-scale datasets of high-resolution natural images.

\begin{figure}[!t]
\renewcommand{\baselinestretch}{1.0}
\centering
\includegraphics[width=0.23\textwidth]{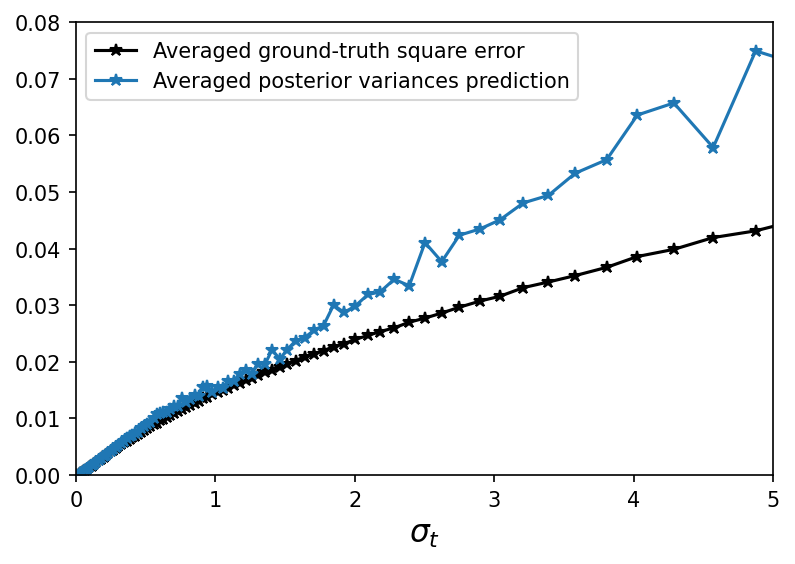}
\includegraphics[width=0.23\textwidth]{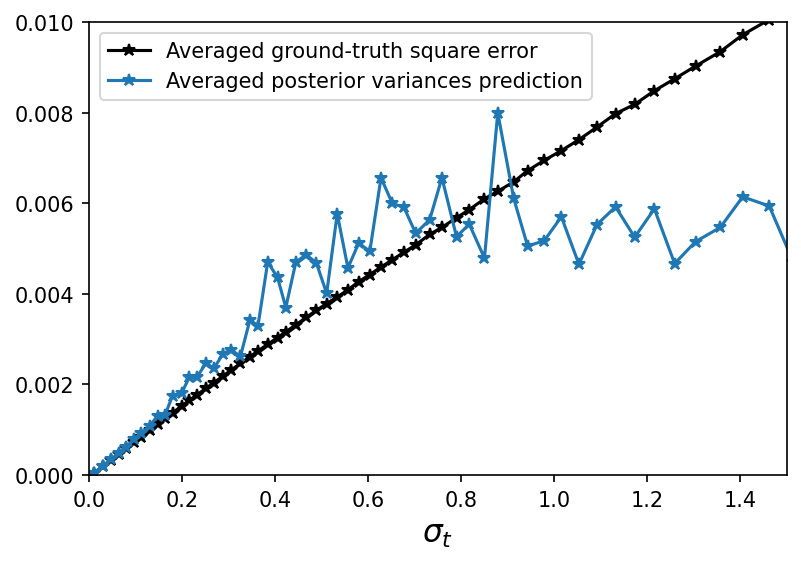}
\caption{\textbf{Averaged values of $\mathbf{e}$ (black line) and $\mathbf{r}_t^2(\mathbf{x}_t)$ (blue line)} on FFHQ (Left) and ImageNet (Right).}\label{fig:plot-posvar}
\end{figure}
\begin{figure}[!t]
\centering
\includegraphics[width=0.48\textwidth]{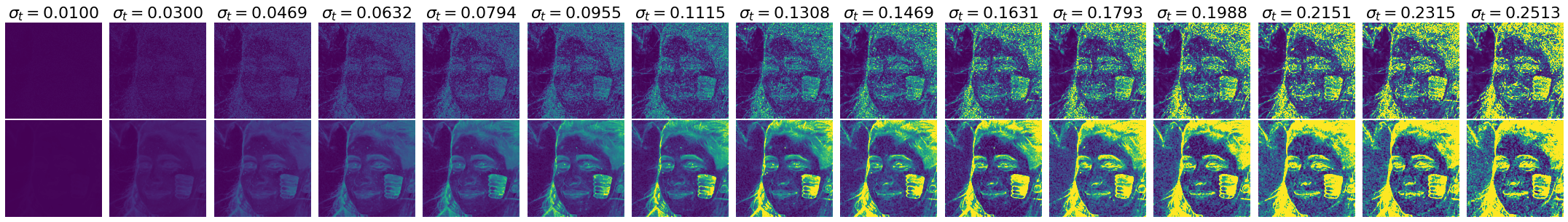}
\caption{\textbf{Visualization of $\mathbf{e}$ and $\hat{\mathbf{r}}_t^2(\mathbf{x}_t)$ of an example image $\mathbf{x}_0$ at different $t$.} Top: $\mathbf{e}$; Bottom: $\hat{\mathbf{r}}_t^2(\mathbf{x}_t)$. The results are averaged over RGB channels for better visualization.}\label{fig:viz-posvar}
\end{figure}

\begin{table*}[!t]
\renewcommand{\baselinestretch}{1.0}
\renewcommand{\arraystretch}{1.0}
\setlength{\tabcolsep}{2.2pt}
\centering
\small
\begin{tabular}{@{}cccccccccccccc@{}}
\toprule
\multirow{2}{*}{\textbf{Dataset}} & \multirow{2}{*}{\textbf{Method}}      & \multicolumn{3}{c}{\textbf{Inpaint~(Random)}} & \multicolumn{3}{c}{\textbf{Deblur~(Gaussian)}} & \multicolumn{3}{c}{\textbf{Deblur~(Motion)}} & \multicolumn{3}{c}{\textbf{Super resolution~($4\times$)}} \\
\cmidrule(lr){3-5}  \cmidrule(lr){6-8}  \cmidrule(lr){9-11}  \cmidrule(lr){12-14} 
&     & SSIM~$\uparrow$ & LPIPS~$\downarrow$ & FID~$\downarrow$ & SSIM~$\uparrow$ & LPIPS~$\downarrow$ & FID~$\downarrow$ & SSIM~$\uparrow$ & LPIPS~$\downarrow$ & FID~$\downarrow$ & SSIM~$\uparrow$ & LPIPS~$\downarrow$ & FID~$\downarrow$ \\
\midrule
\multirow{6}{*}{FFHQ}      & Convert~(\textit{Ours})   & \textbf{0.9279}   & \textbf{0.0794}    & \textbf{25.90}    & 0.7905 & \textbf{0.1836} & \textbf{52.42} & \underline{0.7584} & \underline{0.2156} & \textbf{62.88} & \underline{0.7878} & \underline{0.1962} & \underline{58.37}      \\
                           & Analytic~(\textit{Ours})  & \underline{0.9272} & \underline{0.0845} & 28.83 & \underline{0.7926} & 0.1850 & \underline{53.09}  & 0.7579 & 0.2183 & \underline{64.77} & \underline{0.7878} & 0.1968 & 59.83\\
                           & DWT-Var~(\textit{Ours})       & 0.9209 & 0.0863 & \underline{28.54} & \textbf{0.7968} & \underline{0.1837} & 57.52 & \textbf{0.7677} & \textbf{0.2103} & 65.34 & \textbf{0.8025} & \textbf{0.1856} & \textbf{57.26} \\
                           \cmidrule(lr){2-14}
                          & TMPD                      & 0.8224 & 0.1924             & 70.92             & 0.7289 & 0.2523 & 76.52  & 0.7014 & 0.2718 & 83.49  & 0.7085 & 0.2701 & 79.58 \\
                          & DPS                       & 0.8891 & 0.1323             & 49.46             & 0.6284 & 0.3652 & 136.12 & 0.4904 & 0.4924 & 212.48 & 0.7719 & 0.2054 & 61.36 \\
                          & $\Pi\text{GDM}$           & 0.8784 & 0.1422             & 49.89             & 0.7890 & 0.1910 & 59.93  & 0.7543 & 0.2209 & 66.14  & 0.7850 & 0.2005 & 61.46 \\
\midrule
\midrule
\multirow{5}{*}{ImageNet} & Convert~(\textit{Ours})   & \underline{0.8559} & \textbf{0.1329}    & \textbf{29.14}    & \underline{0.6007} & \textbf{0.3327}      & \underline{95.23}  & \textbf{0.5634}    & \textbf{0.3656}    & \textbf{109.61} & 0.5869 & \underline{0.3477} & 96.76 \\
                          & Analytic~(\textit{Ours})  & 0.8481             & \underline{0.1446} & \underline{35.51} & \textbf{0.6009}    & \underline{0.3334}   & \textbf{93.21}     & \underline{0.5611} & \underline{0.3668} & \underline{113.39}  & \textbf{0.5958} & 0.3495 & \underline{95.33} \\
                          \cmidrule(lr){2-14}
                          & TMPD                      & 0.7011             & 0.2892             & 293.80            & 0.5430             & 0.4114               & 291.29             & 0.4773             & 0.4567 & 302.40 & 0.5186 & 0.4298 & 296.73 \\
                          & DPS                       & \textbf{0.8623}    & 0.1490             & 36.58             & 0.4603             & 0.4630               & 173.77             & 0.3582             & 0.5554 & 282.21 & 0.5860 & \textbf{0.3231} & \textbf{92.89} \\
                          & $\Pi\text{GDM}$           & 0.7658             & 0.2328             & 64.96             & 0.5946             & 0.3429               & 102.89             & 0.5534             & 0.3781 & 113.89 & \underline{0.5925} & 0.3552 & 100.36 \\
\bottomrule
\end{tabular}
\caption{\textbf{Quantitative results on FFHQ and ImageNet dataset for Type I guidance.} We use \textbf{bold} and \underline{underline} for the best and second best, respectively. Note that DPS here uses the step size of $1/ (2\sigma^2)$ for pure covariance comparisons.}\label{tab:typeIquant}
\end{table*}
\begin{figure*}
\centering
\includegraphics[width=1\textwidth]{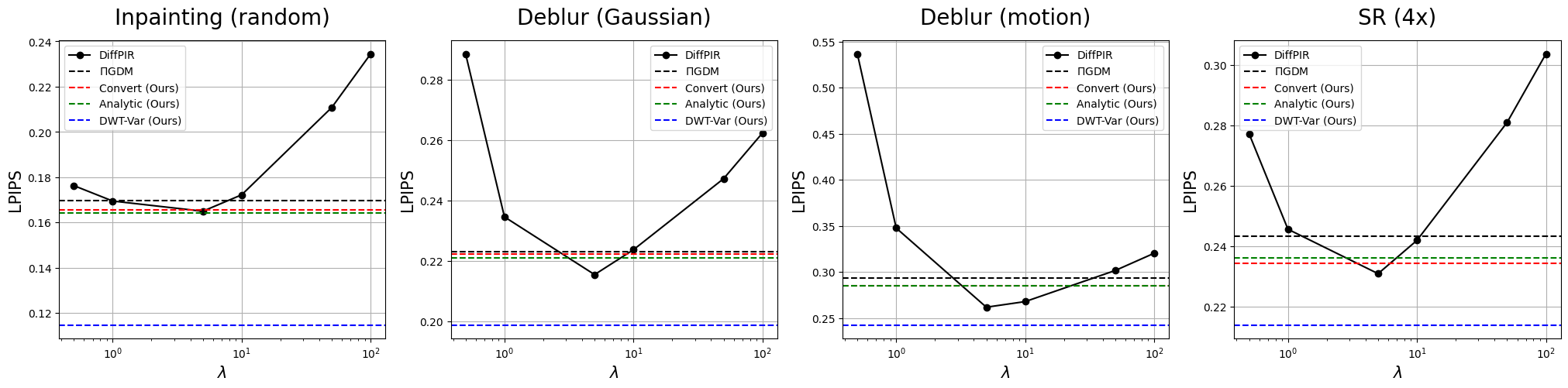}
\caption{\textbf{LPIPS comparisons on FFHQ for Type II guidance.} For DiffPIR, we report LPIPS under different $\lambda$.}\label{fig:typeIIquant}
\end{figure*}

\section{Experiments}
In this section, we evaluate the proposed methods on a range of inverse problems, including inpainting, debluring, and super resolution. To minimize the effect of different implementations for fair comparison, our techniques and prior works are all implemented on a unified codebase based on an open-source diffusion codebase k-diffusion\footnote{\url{https://github.com/crowsonkb/k-diffusion}}. Additional details and results are included in Appendix~\ref{sec:add-exp}. The source code is available at \url{https://github.com/xypeng9903/k-diffusion-inverse-problems}.



\subsection{Sanity Check for Converting Reverse Variance}
The derivation of Eq.~\eqref{eq:convert} is based on the perfect model. However, estimation error of reverse variance in practice may lead to failure when applying Eq.~\eqref{eq:convert}. To validate the effectiveness of Eq.~\eqref{eq:convert} in practice, we compare the ground-truth square errors made by the denoiser $\mathbf{e} = (\mathbf{x}_0 - D_t(\mathbf{x}_t))^2$ with the posterior variance prediction $\hat{\mathbf{r}}_t^2(\mathbf{x}_t)$. We compare $\mathbf{e}$ with $\hat{\mathbf{r}}_t^2(\mathbf{x}_t)$ since the posterior variance $\mathbf{r}_t^{*2}(\mathbf{x}_t)$ is the MMSE estimator of $\mathbf{e}$ given $\mathbf{x}_t$ (assuming $D_t(\mathbf{x}_t) = \mathbb{E}[\mathbf{x}_0|\mathbf{x}_t]$), as suggested by Eq.~\eqref{eq:opt-pos-var}. This implies that a good $\hat{\mathbf{r}}_t^2(\mathbf{x}_t)$ should be a reliable predictor of $\mathbf{e}$. Thus, we compare $\mathbf{e}$ with $\hat{\mathbf{r}}_t^2(\mathbf{x}_t)$ as a sanity check to gauge the effectiveness of Eq.~\eqref{eq:convert} in practice. In Figure~\ref{fig:plot-posvar}, we plot their averaged values over all the pixels and test images. The posterior variances prediction obtained via Eq.~\eqref{eq:convert} is accurate only in the regions with low noise level. This is reasonable since the optimal reverse variance $\mathbf{v}^{*2}(\mathbf{x}_t)$ is bounded by the upper bound $\beta_t$ and the lower bound $\Tilde{\beta}_t$. The two bounds are almost equal in the regions with high noise level~\cite{nichol2021improved}. When $t$ is large, $\mathbf{v}^{*2}(\mathbf{x}_t)-\Tilde{\beta}_t \approx \mathbf{0}$ and Eq.~\eqref{eq:convert} becomes a $0/0$ limit that possesses high numerical instability. In Figure~\ref{fig:viz-posvar}, we visualize $\mathbf{e}$ and $\hat{\mathbf{r}}_t^2(\mathbf{x}_t)$ of an image $\mathbf{x}_0$ at different $t$ and show that $\mathbf{e}$ can be well predicted by $\hat{\mathbf{r}}_t^2(\mathbf{x}_t)$ when $t$ is small.

\begin{table*}[!t]
\renewcommand{\baselinestretch}{1.0}
\renewcommand{\arraystretch}{1.0}
\setlength{\tabcolsep}{2.4pt}
\centering
\small
\begin{tabular}{@{}cccccccccccccc@{}}
\toprule
        \multirow{2}{*}{\textbf{Dataset}} & \multirow{2}{*}{\textbf{Method}}      & \multicolumn{3}{c}{\textbf{Inpaint~(Random)}} & \multicolumn{3}{c}{\textbf{Deblur~(Gaussian)}} & \multicolumn{3}{c}{\textbf{Deblur~(Motion)}} & \multicolumn{3}{c}{\textbf{Super resolution~($4\times$)}} \\
        \cmidrule(lr){3-5}  \cmidrule(lr){6-8}  \cmidrule(lr){9-11}  \cmidrule(lr){12-14} 
                                          &     & SSIM~$\uparrow$ & LPIPS~$\downarrow$ & FID~$\downarrow$ & SSIM~$\uparrow$ & LPIPS~$\downarrow$ & FID~$\downarrow$ & SSIM~$\uparrow$ & LPIPS~$\downarrow$ & FID~$\downarrow$ & SSIM~$\uparrow$ & LPIPS~$\downarrow$ & FID~$\downarrow$ \\
        \midrule
        \multirow{3}{*}{FFHQ}                               
                                  & Convert~(\textit{Ours})   & \textbf{0.9241} & \textbf{0.0822}    & \textbf{27.50}    & \underline{0.7783} & \textbf{0.1969} & \textbf{59.31} & \underline{0.7329} & \textbf{0.2324} & \textbf{66.18} & \textbf{0.7632} & \textbf{0.2183} & \textbf{67.17} \\
                                  & Analytic~(\textit{Ours})  & \underline{0.9232} & \underline{0.0852} & \underline{28.63} & \textbf{0.7790} & \underline{0.1971} & \underline{59.80} & \textbf{0.7331} & \underline{0.2336} & \underline{69.82} & \underline{0.7622} & \underline{0.2195} & \underline{68.84}                     \\
                                  & $\Pi\text{GDM}$           & 0.7078 & 0.2605 & 77.46 & 0.7221 & 0.2421 & 71.19 & 0.6977 & 0.2607 & 75.15 & 0.7205 & 0.2442 & 72.41   \\                
        \midrule 
        \midrule
        \multirow{3}{*}{ImageNet} 
                                  & Convert~(\textit{Ours})   & \textbf{0.8492} & \textbf{0.1394}    & \textbf{33.46} & \textbf{0.5770} & \textbf{0.3568} & \textbf{100.50} & \textbf{0.5341} & \textbf{0.3944} & \underline{131.48} & \textbf{0.5613} & \textbf{0.3846} & \textbf{116.27} \\
                                  & Analytic~(\textit{Ours})  & \underline{0.8417} & \underline{0.1484} & \underline{37.98} & \underline{0.5722} & \underline{0.3583} & \underline{103.67} & \textbf{0.5341} & \underline{0.3947} & \textbf{126.39} & \underline{0.5537} & \underline{0.3867} & \underline{116.87} \\
                                  & $\Pi\text{GDM}$           & 0.5102 & 0.4293 & 141.80 & 0.5071 & 0.4095 & 127.41 & 0.4887 & 0.4267 & 132.38 & 0.5150 & 0.4098 & 127.40 \\
\bottomrule
\end{tabular}
\caption{\textbf{Results for $\Pi\text{GDM}$ with adaptive weight on FFHQ and ImageNet dataset.} We use \textbf{bold} and \underline{underline} for the best and second best, respectively.}\label{tab:complete-pgdm}
\end{table*}
\begin{figure*}
\centering
\includegraphics[width=1\textwidth]{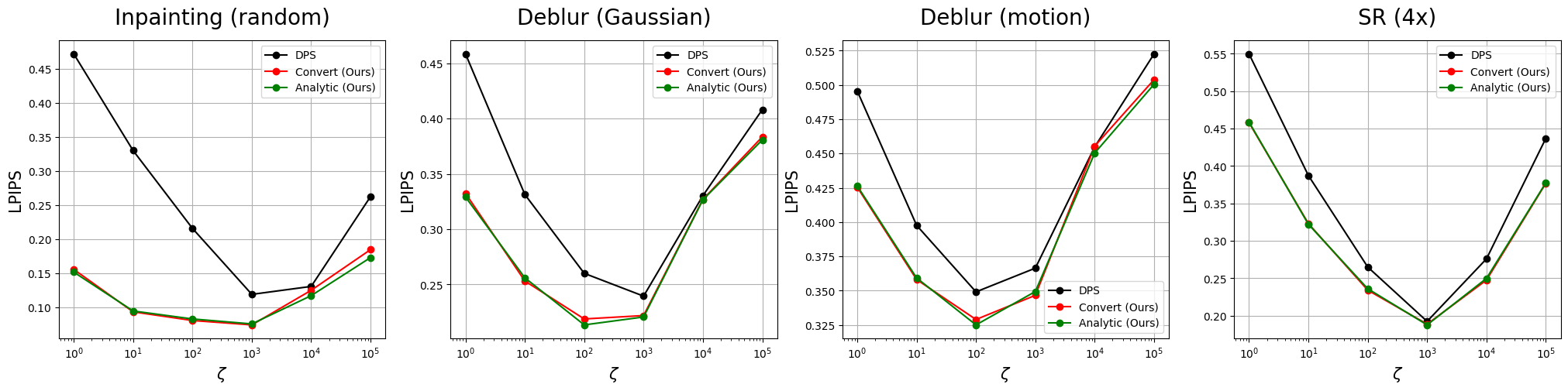}
\caption{\textbf{LPIPS comparisons on FFHQ for DPS with heuristic step size $\zeta_t = \zeta / \lVert \mathbf{y} - \mathbf{A}D_t(\mathbf{x}_t)\rVert_2$.} For comprehensive comparisons, we report LPIPS under different $\zeta$.}\label{fig:complete-dps-lpips}
\end{figure*}

\subsection{Quantitative Results}\label{sec:main-quant}
\textbf{Experimental setup.} Following~\cite{chung2023diffusion, wang2023zeroshot}, we perform experiments on the FFHQ 256$\times$256 and ImageNet 256$\times$256 datasets to compare different methods with unconditional diffusion models from \cite{chung2023diffusion} and \cite{dhariwal2021diffusion}, respectively. We learn optimal variance in DWT domain on FFHQ dataset by modifying the FFHQ model (see Appendix~\ref{app:ot} for details). The degradation models are specified mostly following~\cite{zhu2023denoising}: (i) For inpainting, we randomly mask 50 percent of the total pixels. (ii) For Gaussian debluring and motion debluring, we use the same setup of the bluring kernels to~\cite{chung2023diffusion}. (iii) For super resolution~(SR), we consider bicubic downsampling. All measurements are corrupted by Gaussian noise with $\sigma=0.05$. To evaluate different methods, we follow \citet{chung2023diffusion} to use three metrics: Structure Similarity Index Measure~(SSIM), Learned Perceptual Image Patch Similarity~(LPIPS, \citet{zhang2018unreasonable}) and Frechet Inception Distance~(FID, \citet{heusel2017gans}). 
For the sampler setup, all Type I methods use the same Heun's 2nd deterministic sampler suggested in~\citet{karras2022elucidating} with 50 sampling steps, and all Type II methods use the same Heun’s 2nd stochastic sampler~($S_{\text{churn}}=80, S_{\text{tmin}}=0.05, S_{\text{tmax}}=50, S_{\text{noise}}=1.003$, definition see~\citet{karras2022elucidating}) with 50 sampling steps since we found that Type II methods does not perform well using deterministic samplers\footnote{To understand how to leverage pre-trained DDPM model to perform sampling under perturbation kernels given in Section~\ref{sec:intro dpm}, please refer to Appendix~\ref{app:convert kernels}.}. 

In initial experiments, we observed that using the spatial variance predicted by the proposed plug-and-play methods (Section~\ref{sec:convert} and Section~\ref{sec:analytic}) for all sampling steps results in poor performance. This can be attributed to diagonal Gaussian variational $q_t(\mathbf{x}_0|\mathbf{x}_t)$ being an effective approximation to posterior $p_t(\mathbf{x}_0|\mathbf{x}_t)$ only for small noise level~\cite{sohl2015deep, xiao2022tackling}. To address this, we heuristically use the spatial variance only at the last several sampling steps with low noise level but use $\Pi\text{GDM}$ covariance for high noise level. We empirically find that sampling with spatial variance yields highly stable results when $\sigma_t<0.2$~(\emph{i.e.}, 12 out of 50 steps). Nevertheless, we observed that when using DWT variance (Section~\ref{sec:ot}), the posterior approximation is accurate enough such that the variance can be used for all sampling steps. In the experiments, DWT variance is used when $\sigma_t < 1$ for efficiency, otherwise many CG steps are required to compute.

\begin{figure*}[!t]
\centering
\includegraphics[width=1\textwidth]{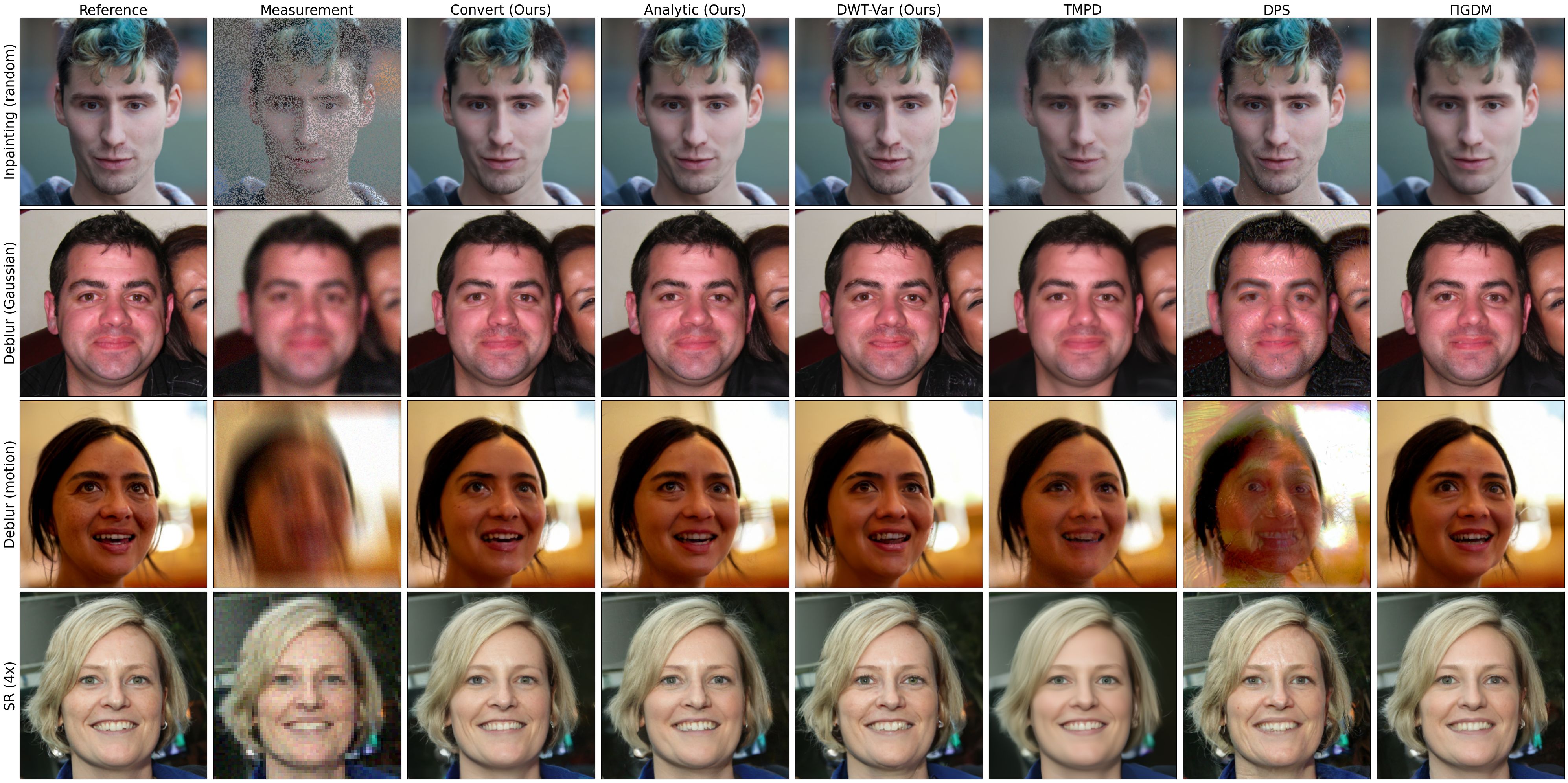}
\caption{\textbf{Qualitative results for Table~\ref{tab:typeIquant} on FFHQ dataset.} We observed that our methods reconstruct fine details of the image more faithfully compared to baselines.}\label{fig:viz}
\end{figure*}

\textbf{Results for pure covariance comparisons.} Table~\ref{tab:typeIquant} and Figure~\ref{fig:typeIIquant} summarize the results for Type I and Type II guidance. \texttt{DPS, $\Pi\text{GDM}$, DiffPIR} refer to the posterior covariance in Table~\ref{tab:compare}; \texttt{Convert}, \texttt{Analytic}, \texttt{DWT-Var}, and \texttt{TMPD} refer to the posterior covariance obtained using approaches presented in Sections~\ref{sec:convert},~\ref{sec:analytic},~\ref{sec:ot}, and $\sigma_t^2\mathrm{diag}[\nabla_{\mathbf{x}_t}\mathbf{1}^TD_t(\mathbf{x}_t)]$ used in~\cite{boys2023tweedie}, respectively. Note that for Type I guidance, our methods achieve the best results in almost all tasks. Although DPS outperforms us in several cases, we observe that its performance is very unstable~(see DPS performance in debluring tasks). In Figure~\ref{fig:viz}, we visualiza reconstruction results of different methods. We observed that  our methods reconstruct fine details of the image more
faithfully compared to baselines. For Type II guidance, the plug-and-play methods obtain comparable performance to optimally-tuned DiffPIR and \texttt{DWT-Var} outperforms optimally-tuned DiffPIR by a significant margin without any hyper-parameter tuning.

\textbf{Results with heuristics implementations.} In addition to using Gaussian for denoising posterior approximation, prior works also propose several heuristics. Here we show that the methods with heuristics can also be improved by replacing the last sampling steps with our methods. In particular, $\Pi\text{GDM}$ introduces an \textit{adaptive weight} to the likelihood score $\nabla_{\mathbf{x}_t} \log p_t(\mathbf{y}|\mathbf{x}_t)$ to adjust the guidance strength according to the timestep, which rescales the likelihood score by $r_t^2$ with $r_t^2 = \sigma_t^2 / (1 + \sigma_t^2)$. Table~\ref{tab:complete-pgdm} summarizes the results for $\Pi\text{GDM}$ with adaptive weight. Obviously, $\Pi\text{GDM}$ achieves significant performance gain using Type I guidance with optimal variances at the last sampling steps.
As for DPS, it introduces a trick to heuristically determine the step size of guidance, as mentioned in Section~\ref{sec:tIguid}. Since this trick involves a hyper-parameter $\zeta$, we report the performance under different $\zeta$. Figure~\ref{fig:complete-dps-lpips} summarizes the quantitative results for DPS with heuristic guidance strength. DPS with the last sampling steps replaced by our methods can achieve the best results across all hyper-parameters. 

\section{Conclusions}
We reveal that recent methods can be uniformly interpreted as approximating the conditional posterior mean by employing Gaussian distribution with hand-crafted isotropic covariance to approximate the intractable denoising posterior. Inspired by this, we propose to enhance existing methods by using more principled covariance determined by maximum likelihood estimation. Experimental results show that the proposed methods significantly outperform existing methods and also eliminate the need for hyperparameter tuning.

\textbf{Limitations.} Despite the reasonable improvements our approaches offer over prior works, it is worth noting that, due to the diagonal constraint, the proposed covariance cannot match the optimum exactly even when the model is perfectly trained. Future research could explore better principles for designing the covariance~\cite{dorta2018structured, nehme2024uncertainty}, as well as leveraging nonlinear transformation~\cite{balle2020nonlinear,zhu2022transformerbased,li2024frequencyaware} for better correlation reduction. Additionally, methods like efficient approximation for Tweedie's approach could be investigated. These represent intriguing directions for future work.

\section*{Acknowledgement}
This work was supported in part by the National Natural Science Foundation of China under Grant 62125109, Grant 61931023, Grant 61932022, Grant 62371288, Grant 62320106003, Grant 62301299, Grant T2122024, Grant 62120106007, Grant 62250055.

\section*{Impact Statement}
This paper presents work whose goal is to advance the field of Machine Learning. There are many potential societal consequences of our work, none which we feel must be specifically highlighted here.


\begin{thebibliography}{43}
\providecommand{\natexlab}[1]{#1}
\providecommand{\url}[1]{\texttt{#1}}
\expandafter\ifx\csname urlstyle\endcsname\relax
  \providecommand{\doi}[1]{doi: #1}\else
  \providecommand{\doi}{doi: \begingroup \urlstyle{rm}\Url}\fi

\bibitem[Ball{\'e} et~al.(2020)Ball{\'e}, Chou, Minnen, Singh, Johnston,
  Agustsson, Hwang, and Toderici]{balle2020nonlinear}
Ball{\'e}, J., Chou, P.~A., Minnen, D., Singh, S., Johnston, N., Agustsson, E.,
  Hwang, S.~J., and Toderici, G.
\newblock Nonlinear transform coding.
\newblock \emph{IEEE Journal of Selected Topics in Signal Processing},
  15\penalty0 (2):\penalty0 339--353, 2020.

\bibitem[Bao et~al.(2022{\natexlab{a}})Bao, Li, Sun, Zhu, and
  Zhang]{pmlr-v162-bao22d}
Bao, F., Li, C., Sun, J., Zhu, J., and Zhang, B.
\newblock Estimating the optimal covariance with imperfect mean in diffusion
  probabilistic models.
\newblock In \emph{Proceedings of the 39th International Conference on Machine
  Learning}, pp.\  1555--1584, 2022{\natexlab{a}}.

\bibitem[Bao et~al.(2022{\natexlab{b}})Bao, Li, Zhu, and
  Zhang]{bao2022analyticdpm}
Bao, F., Li, C., Zhu, J., and Zhang, B.
\newblock Analytic-{DPM}: an analytic estimate of the optimal reverse variance
  in diffusion probabilistic models.
\newblock In \emph{The 10th International Conference on Learning
  Representations}, 2022{\natexlab{b}}.
\newblock URL \url{https://openreview.net/forum?id=0xiJLKH-ufZ}.

\bibitem[Bishop(2006)]{bishop2006pattern}
Bishop, C.~M.
\newblock \emph{Pattern Recognition and Machine Learning}.
\newblock Springer, 2006.

\bibitem[Boys et~al.(2023)Boys, Girolami, Pidstrigach, Reich, Mosca, and
  Akyildiz]{boys2023tweedie}
Boys, B., Girolami, M., Pidstrigach, J., Reich, S., Mosca, A., and Akyildiz,
  O.~D.
\newblock Tweedie moment projected diffusions for inverse problems.
\newblock \emph{arXiv preprint arXiv:2310.06721}, 2023.

\bibitem[Chan et~al.(2023)Chan, Young, and Metzler]{chan2023sud}
Chan, M.~A., Young, S.~I., and Metzler, C.~A.
\newblock {SUD$^2$}: Supervision by denoising diffusion models for image
  reconstruction.
\newblock In \emph{NeurIPS 2023 Deep Inverse Workshop}, 2023.

\bibitem[Choi et~al.(2021)Choi, Kim, Jeong, Gwon, and Yoon]{choi2021ilvr}
Choi, J., Kim, S., Jeong, Y., Gwon, Y., and Yoon, S.
\newblock {ILVR}: Conditioning method for denoising diffusion probabilistic
  models.
\newblock In \emph{Proceedings of the IEEE/CVF International Conference on
  Computer Vision}, pp.\  14347--14356, 2021.

\bibitem[Chung et~al.(2023{\natexlab{a}})Chung, Kim, Mccann, Klasky, and
  Ye]{chung2023diffusion}
Chung, H., Kim, J., Mccann, M.~T., Klasky, M.~L., and Ye, J.~C.
\newblock Diffusion posterior sampling for general noisy inverse problems.
\newblock In \emph{The 11th International Conference on Learning
  Representations}, 2023{\natexlab{a}}.
\newblock URL \url{https://openreview.net/forum?id=OnD9zGAGT0k}.

\bibitem[Chung et~al.(2023{\natexlab{b}})Chung, Lee, and Ye]{chung2023fast}
Chung, H., Lee, S., and Ye, J.~C.
\newblock Fast diffusion sampler for inverse problems by geometric
  decomposition.
\newblock \emph{arXiv preprint arXiv:2303.05754}, 2023{\natexlab{b}}.

\bibitem[Dhariwal \& Nichol(2021)Dhariwal and Nichol]{dhariwal2021diffusion}
Dhariwal, P. and Nichol, A.
\newblock Diffusion models beat {GAN}s on image synthesis.
\newblock In \emph{Advances in Neural Information Processing Systems 34}, pp.\
  8780--8794, 2021.

\bibitem[Dorta et~al.(2018)Dorta, Vicente, Agapito, Campbell, and
  Simpson]{dorta2018structured}
Dorta, G., Vicente, S., Agapito, L., Campbell, N.~D., and Simpson, I.
\newblock Structured uncertainty prediction networks.
\newblock In \emph{Proceedings of the IEEE Conference on Computer Vision and
  Pattern Recognition}, pp.\  5477--5485, 2018.

\bibitem[Feng et~al.(2023)Feng, Smith, Rubinstein, Chang, Bouman, and
  Freeman]{RN320}
Feng, B.~T., Smith, J., Rubinstein, M., Chang, H., Bouman, K.~L., and Freeman,
  W.~T.
\newblock Score-based diffusion models as principled priors for inverse
  imaging.
\newblock In \emph{Proceedings of the IEEE/CVF International Conference on
  Computer Vision}, pp.\  10520--10531, 2023.

\bibitem[Goyal(2001)]{goyal2001theoretical}
Goyal, V.~K.
\newblock Theoretical foundations of transform coding.
\newblock \emph{IEEE Signal Processing Magazine}, 18\penalty0 (5):\penalty0
  9--21, 2001.

\bibitem[Heusel et~al.(2017)Heusel, Ramsauer, Unterthiner, Nessler, and
  Hochreiter]{heusel2017gans}
Heusel, M., Ramsauer, H., Unterthiner, T., Nessler, B., and Hochreiter, S.
\newblock {GANs} trained by a two time-scale update rule converge to a local
  {Nash} equilibrium.
\newblock In \emph{Advances in Neural Information Processing Systems 30}, pp.\
  6626--6637, 2017.

\bibitem[Ho et~al.(2020)Ho, Jain, and Abbeel]{ho2020denoising}
Ho, J., Jain, A., and Abbeel, P.
\newblock Denoising diffusion probabilistic models.
\newblock In \emph{Advances in Neural Information Processing Systems 33}, pp.\
  6840--6851, 2020.

\bibitem[Karras et~al.(2022)Karras, Aittala, Aila, and
  Laine]{karras2022elucidating}
Karras, T., Aittala, M., Aila, T., and Laine, S.
\newblock Elucidating the design space of diffusion-based generative models.
\newblock In \emph{Advances in Neural Information Processing Systems 35}, pp.\
  26565--26577, 2022.

\bibitem[Kingma et~al.(2021)Kingma, Salimans, Poole, and
  Ho]{kingma2021variational}
Kingma, D., Salimans, T., Poole, B., and Ho, J.
\newblock Variational diffusion models.
\newblock In \emph{Advances in Neural Information Processing Systems 34}, pp.\
  21696--21707, 2021.

\bibitem[Li et~al.(2024)Li, Li, Dai, Li, Zou, and Xiong]{li2024frequencyaware}
Li, H., Li, S., Dai, W., Li, C., Zou, J., and Xiong, H.
\newblock Frequency-aware transformer for learned image compression.
\newblock In \emph{The 12th International Conference on Learning
  Representations}, 2024.
\newblock URL \url{https://openreview.net/forum?id=HKGQDDTuvZ}.

\bibitem[Lu et~al.(2022)Lu, Zheng, Bao, Chen, Li, and Zhu]{lu2022maximum}
Lu, C., Zheng, K., Bao, F., Chen, J., Li, C., and Zhu, J.
\newblock Maximum likelihood training for score-based diffusion odes by high
  order denoising score matching.
\newblock In \emph{Proceedings of the 39th International Conference on Machine
  Learning}, pp.\  14429--14460, 2022.

\bibitem[Lugmayr et~al.(2022)Lugmayr, Danelljan, Romero, Yu, Timofte, and
  Van~Gool]{lugmayr2022repaint}
Lugmayr, A., Danelljan, M., Romero, A., Yu, F., Timofte, R., and Van~Gool, L.
\newblock Repaint: Inpainting using denoising diffusion probabilistic models.
\newblock In \emph{Proceedings of the IEEE/CVF Conference on Computer Vision
  and Pattern Recognition}, pp.\  11461--11471, 2022.

\bibitem[Luo et~al.(2023)Luo, Gustafsson, Zhao, Sj{\"o}lund, and
  Sch{\"o}n]{luo2023refusion}
Luo, Z., Gustafsson, F.~K., Zhao, Z., Sj{\"o}lund, J., and Sch{\"o}n, T.~B.
\newblock Refusion: Enabling large-size realistic image restoration with
  latent-space diffusion models.
\newblock In \emph{Proceedings of the IEEE/CVF Conference on Computer Vision
  and Pattern Recognition}, pp.\  1680--1691, 2023.

\bibitem[Mardani et~al.(2024)Mardani, Song, Kautz, and
  Vahdat]{mardani2023variational}
Mardani, M., Song, J., Kautz, J., and Vahdat, A.
\newblock A variational perspective on solving inverse problems with diffusion
  models.
\newblock In \emph{The 12th International Conference on Learning
  Representations}, 2024.

\bibitem[Meng et~al.(2021)Meng, Song, Li, and Ermon]{meng2021estimating}
Meng, C., Song, Y., Li, W., and Ermon, S.
\newblock Estimating high order gradients of the data distribution by
  denoising.
\newblock In \emph{Advances in Neural Information Processing Systems 34}, pp.\
  25359--25369, 2021.

\bibitem[Nehme et~al.(2023)Nehme, Yair, and Michaeli]{nehme2024uncertainty}
Nehme, E., Yair, O., and Michaeli, T.
\newblock Uncertainty quantification via neural posterior principal components.
\newblock In \emph{Advances in Neural Information Processing Systems 36}, pp.\
  37128--37141, 2023.

\bibitem[Nichol \& Dhariwal(2021)Nichol and Dhariwal]{nichol2021improved}
Nichol, A.~Q. and Dhariwal, P.
\newblock Improved denoising diffusion probabilistic models.
\newblock In \emph{Proceedings of the 38th International Conference on Machine
  Learning}, pp.\  8162--8171, 2021.

\bibitem[Pokle et~al.(2023)Pokle, Muckley, Chen, and Karrer]{pokle2023training}
Pokle, A., Muckley, M.~J., Chen, R.~T., and Karrer, B.
\newblock Training-free linear image inversion via flows.
\newblock \emph{arXiv preprint arXiv:2310.04432}, 2023.

\bibitem[Ravula et~al.(2023)Ravula, Levac, Jalal, Tamir, and
  Dimakis]{ravula2023optimizing}
Ravula, S., Levac, B., Jalal, A., Tamir, J.~I., and Dimakis, A.~G.
\newblock Optimizing sampling patterns for compressed sensing {MRI} with
  diffusion generative models.
\newblock \emph{arXiv preprint arXiv:2306.03284}, 2023.

\bibitem[Rezende \& Viola(2018)Rezende and Viola]{rezende2018taming}
Rezende, D.~J. and Viola, F.
\newblock Taming {VAE}s.
\newblock \emph{arXiv preprint arXiv:1810.00597}, 2018.

\bibitem[Rout et~al.(2023)Rout, Chen, Kumar, Caramanis, Shakkottai, and
  Chu]{rout2023beyond}
Rout, L., Chen, Y., Kumar, A., Caramanis, C., Shakkottai, S., and Chu, W.-S.
\newblock Beyond first-order {Tweedie}: Solving inverse problems using latent
  diffusion.
\newblock \emph{arXiv preprint arXiv:2312.00852}, 2023.

\bibitem[Saharia et~al.(2022)Saharia, Ho, Chan, Salimans, Fleet, and
  Norouzi]{saharia2022image}
Saharia, C., Ho, J., Chan, W., Salimans, T., Fleet, D.~J., and Norouzi, M.
\newblock Image super-resolution via iterative refinement.
\newblock \emph{IEEE Transactions on Pattern Analysis and Machine
  Intelligence}, 45\penalty0 (4):\penalty0 4713--4726, 2022.

\bibitem[Sohl-Dickstein et~al.(2015)Sohl-Dickstein, Weiss, Maheswaranathan, and
  Ganguli]{sohl2015deep}
Sohl-Dickstein, J., Weiss, E., Maheswaranathan, N., and Ganguli, S.
\newblock Deep unsupervised learning using nonequilibrium thermodynamics.
\newblock In \emph{Proceeding of the 35th International Conference on Machine
  Learning}, pp.\  2256--2265, 2015.

\bibitem[Song et~al.(2021{\natexlab{a}})Song, Meng, and
  Ermon]{song2021denoising}
Song, J., Meng, C., and Ermon, S.
\newblock Denoising diffusion implicit models.
\newblock In \emph{The 9th International Conference on Learning
  Representations}, 2021{\natexlab{a}}.
\newblock URL \url{https://openreview.net/forum?id=St1giarCHLP}.

\bibitem[Song et~al.(2023)Song, Vahdat, Mardani, and
  Kautz]{song2023pseudoinverseguided}
Song, J., Vahdat, A., Mardani, M., and Kautz, J.
\newblock Pseudoinverse-guided diffusion models for inverse problems.
\newblock In \emph{The 11th International Conference on Learning
  Representations}, 2023.
\newblock URL \url{https://openreview.net/forum?id=9_gsMA8MRKQ}.

\bibitem[Song \& Ermon(2019)Song and Ermon]{song2019generative}
Song, Y. and Ermon, S.
\newblock Generative modeling by estimating gradients of the data distribution.
\newblock In \emph{Advances in Neural Information Processing Systems 32}, pp.\
  11918--11930, 2019.

\bibitem[Song et~al.(2021{\natexlab{b}})Song, Sohl-Dickstein, Kingma, Kumar,
  Ermon, and Poole]{song2021scorebased}
Song, Y., Sohl-Dickstein, J., Kingma, D.~P., Kumar, A., Ermon, S., and Poole,
  B.
\newblock Score-based generative modeling through stochastic differential
  equations.
\newblock In \emph{The 9th International Conference on Learning
  Representations}, 2021{\natexlab{b}}.
\newblock URL \url{https://openreview.net/forum?id=PxTIG12RRHS}.

\bibitem[Song et~al.(2022)Song, Shen, Xing, and Ermon]{song2022solving}
Song, Y., Shen, L., Xing, L., and Ermon, S.
\newblock Solving inverse problems in medical imaging with score-based
  generative models.
\newblock In \emph{The 10th International Conference on Learning
  Representations}, 2022.
\newblock URL \url{https://openreview.net/forum?id=vaRCHVj0uGI}.

\bibitem[Wang et~al.(2023)Wang, Yu, and Zhang]{wang2023zeroshot}
Wang, Y., Yu, J., and Zhang, J.
\newblock Zero-shot image restoration using denoising diffusion null-space
  model.
\newblock In \emph{The 11th International Conference on Learning
  Representations}, 2023.
\newblock URL \url{https://openreview.net/forum?id=mRieQgMtNTQ}.

\bibitem[Whang et~al.(2022)Whang, Delbracio, Talebi, Saharia, Dimakis, and
  Milanfar]{whang2022deblurring}
Whang, J., Delbracio, M., Talebi, H., Saharia, C., Dimakis, A.~G., and
  Milanfar, P.
\newblock Deblurring via stochastic refinement.
\newblock In \emph{Proceedings of the IEEE/CVF Conference on Computer Vision
  and Pattern Recognition}, pp.\  16293--16303, 2022.

\bibitem[Xiao et~al.(2022)Xiao, Kreis, and Vahdat]{xiao2022tackling}
Xiao, Z., Kreis, K., and Vahdat, A.
\newblock Tackling the generative learning trilemma with denoising diffusion
  {GAN}s.
\newblock In \emph{The 10th International Conference on Learning
  Representations}, 2022.
\newblock URL \url{https://openreview.net/forum?id=JprM0p-q0Co}.

\bibitem[Zhang et~al.(2020)Zhang, Gool, and Timofte]{zhang2020deep}
Zhang, K., Gool, L.~V., and Timofte, R.
\newblock Deep unfolding network for image super-resolution.
\newblock In \emph{Proceedings of the IEEE/CVF Conference on Computer Vision
  and Pattern Recognition}, pp.\  3217--3226, 2020.

\bibitem[Zhang et~al.(2018)Zhang, Isola, Efros, Shechtman, and
  Wang]{zhang2018unreasonable}
Zhang, R., Isola, P., Efros, A.~A., Shechtman, E., and Wang, O.
\newblock The unreasonable effectiveness of deep features as a perceptual
  metric.
\newblock In \emph{Proceedings of the IEEE Conference on Computer Vision and
  Pattern Recognition}, pp.\  586--595, 2018.

\bibitem[Zhu et~al.(2022)Zhu, Yang, and Cohen]{zhu2022transformerbased}
Zhu, Y., Yang, Y., and Cohen, T.
\newblock Transformer-based transform coding.
\newblock In \emph{International Conference on Learning Representations}, 2022.
\newblock URL \url{https://openreview.net/forum?id=IDwN6xjHnK8}.

\bibitem[Zhu et~al.(2023)Zhu, Zhang, Liang, Cao, Wen, Timofte, and
  Van~Gool]{zhu2023denoising}
Zhu, Y., Zhang, K., Liang, J., Cao, J., Wen, B., Timofte, R., and Van~Gool, L.
\newblock Denoising diffusion models for plug-and-play image restoration.
\newblock In \emph{Proceedings of the IEEE/CVF Conference on Computer Vision
  and Pattern Recognition}, pp.\  1219--1229, 2023.

\end{thebibliography}

\newpage
\appendix
\onecolumn

\section{Proofs}

\begin{lemma}[Tweedie's formula] 
\label{lemma:tweedie}
If the joint distribution between $\mathbf{x}_0, \mathbf{x}_t$ is given by $p_t(\mathbf{x}_0, \mathbf{x}_t)=p(\mathbf{x}_0)p_t(\mathbf{x}_t|\mathbf{x}_0)$ with $p_t(\mathbf{x}_t|\mathbf{x}_0)=\mathcal{N}(\mathbf{x}_t|s_t\mathbf{x}_0, s_t^2 \sigma_t^2 \mathbf{I})$ , then $\nabla_{\mathbf{x}_t} \log p_t(\mathbf{x}_t) = \frac{1}{s_t^2 \sigma_t^2}(s_t \mathbb{E}[\mathbf{x}_0|\mathbf{x}_t] - \mathbf{x}_t)$.

\begin{proof}
\begin{align}
    \nabla_{\mathbf{x}_t} \log p_t(\mathbf{x}_t) &= \frac{\nabla_{\mathbf{x}_t} p_t(\mathbf{x}_t)}{p_t(\mathbf{x}_t)} \\
    &= \frac{1}{p_t(\mathbf{x}_t)}\nabla_{\mathbf{x}_t} \int p(\mathbf{x}_0) p_t(\mathbf{x}_t|\mathbf{x}_0) \mathrm{d}\mathbf{x}_0 \\
    &= \frac{1}{p_t(\mathbf{x}_t)} \int p(\mathbf{x}_0) \nabla_{\mathbf{x}_t} p_t(\mathbf{x}_t|\mathbf{x}_0) \mathrm{d}\mathbf{x}_0 \\
    &= \frac{1}{p_t(\mathbf{x}_t)} \int p(\mathbf{x}_0)p_t(\mathbf{x}_t|\mathbf{x}_0) \nabla_{\mathbf{x}_t} \log p_t(\mathbf{x}_t|\mathbf{x}_0) \mathrm{d}\mathbf{x}_0 \\ 
    &= \int p_t(\mathbf{x}_0|\mathbf{x}_t) \nabla_{\mathbf{x}_t} \log p_t(\mathbf{x}_t|\mathbf{x}_0) \mathrm{d}\mathbf{x}_0 \\
    &= \mathbb{E}_{p_t(\mathbf{x}_0|\mathbf{x}_t)}[\nabla_{\mathbf{x}_t} \log p_t(\mathbf{x}_t|\mathbf{x}_0)] \label{eq:logpx}
\end{align}
For Gaussian perturbation kernel $p_t(\mathbf{x}_t|\mathbf{x}_0)=\mathcal{N}(\mathbf{x}_t|s_t\mathbf{x}_0, s_t^2 \sigma_t^2 \mathbf{I})$, we have $\nabla_{\mathbf{x}_t} \log p_t(\mathbf{x}_t|\mathbf{x}_0) = \frac{1}{s_t^2 \sigma_t^2}(s_t \mathbf{x}_0 - \mathbf{x}_t)$. Plug it into Eq.~\eqref{eq:logpx}, we conclude the proof.
\end{proof}
\end{lemma}

\begin{lemma}[Conditional Tweedie's formula] 
\label{lemma:cond-tweedie}
If the joint distribution between $\mathbf{x}_0, \mathbf{y}, \mathbf{x}_t$ is given by $p_t(\mathbf{x}_0,\mathbf{y},\mathbf{x}_t) = p(\mathbf{x}_0)p(\mathbf{y}|\mathbf{x}_0)p_t(\mathbf{x}_t|\mathbf{x}_0)$ with $p_t(\mathbf{x}_t|\mathbf{x}_0)=\mathcal{N}(\mathbf{x}_t|s_t\mathbf{x}_0, s_t^2 \sigma_t^2 \mathbf{I})$ , then $\nabla_{\mathbf{x}_t} \log p_t(\mathbf{x}_t|\mathbf{y}) = \frac{1}{s_t^2 \sigma_t^2}(s_t \mathbb{E}[\mathbf{x}_0|\mathbf{x}_t, \mathbf{y}] - \mathbf{x}_t)$.

\begin{proof}
\begin{align}
    \nabla_{\mathbf{x}_t} \log p_t(\mathbf{x}_t|\mathbf{y}) &= \frac{\nabla_{\mathbf{x}_t} p_t(\mathbf{x}_t|\mathbf{y})}{p_t(\mathbf{x}_t|\mathbf{y})} \\
    &= \frac{1}{p_t(\mathbf{x}_t|\mathbf{y})}\nabla_{\mathbf{x}_t} \int p_t(\mathbf{x}_t|\mathbf{x}_0, \mathbf{y})p(\mathbf{x}_0|\mathbf{y}) \mathrm{d}\mathbf{x}_0 \\
    &= \frac{1}{p_t(\mathbf{x}_t|\mathbf{y})}\nabla_{\mathbf{x}_t} \int p_t(\mathbf{x}_t|\mathbf{x}_0)p(\mathbf{x}_0|\mathbf{y}) \mathrm{d}\mathbf{x}_0 \label{eq:xt-ind-y-giv-x0-1}\\
    &= \frac{1}{p_t(\mathbf{x}_t|\mathbf{y})} \int p(\mathbf{x}_0|\mathbf{y}) \nabla_{\mathbf{x}_t} p_t(\mathbf{x}_t|\mathbf{x}_0) \mathrm{d}\mathbf{x}_0 \\
    &= \frac{1}{p_t(\mathbf{x}_t|\mathbf{y})} \int p(\mathbf{x}_0|\mathbf{y})p_t(\mathbf{x}_t|\mathbf{x}_0, \mathbf{y}) \nabla_{\mathbf{x}_t} \log p_t(\mathbf{x}_t|\mathbf{x}_0) \mathrm{d}\mathbf{x}_0 \label{eq:xt-ind-y-giv-x0-2}\\ 
    &= \int p_t(\mathbf{x}_0|\mathbf{x}_t, \mathbf{y}) \nabla_{\mathbf{x}_t} \log p_t(\mathbf{x}_t|\mathbf{x}_0) \mathrm{d}\mathbf{x}_0 \\
    &= \mathbb{E}_{p_t(\mathbf{x}_0|\mathbf{x}_t, \mathbf{y})}[\nabla_{\mathbf{x}_t} \log p_t(\mathbf{x}_t|\mathbf{x}_0)] \label{eq:logpxy}
\end{align}
where Eq.~\eqref{eq:xt-ind-y-giv-x0-1} and Eq.~\eqref{eq:xt-ind-y-giv-x0-2} are due to the conditional independence between $\mathbf{x}_t$ and $\mathbf{y}$ given $\mathbf{x}_0$, such that $p_t(\mathbf{x}_t|\mathbf{x}_0, \mathbf{y}) = p_t(\mathbf{x}_t|\mathbf{x}_0)$. For Gaussian perturbation kernel $p_t(\mathbf{x}_t|\mathbf{x}_0)=\mathcal{N}(\mathbf{x}_t|s_t\mathbf{x}_0, s_t^2 \sigma_t^2 \mathbf{I})$, we have $\nabla_{\mathbf{x}_t} \log p_t(\mathbf{x}_t|\mathbf{x}_0) = \frac{1}{s_t^2 \sigma_t^2}(s_t \mathbf{x}_0 - \mathbf{x}_t)$. Plug it into Eq.~\eqref{eq:logpxy}, we conclude the proof.
\end{proof}
\end{lemma}

\subsection{Derivation of the Marginal Preserving Property of Diffusion ODEs}
For the sake of completeness, here we prove that the ODEs given in Eq.~\eqref{eq:uncond samp} and Eq.~\eqref{eq: cond sample} respectively maintain the exact same marginals to $p_t(\mathbf{x}_t)$ and $p_t(\mathbf{x}_t|\mathbf{y})$.
\begin{proof}
    By borrowing the results from (Equation 4, \cite{karras2022elucidating}) and setting $s_t = 1, \sigma_t = t$, $\mathbf{x}_t$ determined by the following ODE preserves the marginal $p_t(\mathbf{x}_t)$ for all $t\in [0, T]$:
    \begin{equation}
        \mathrm{d}\mathbf{x}_t = -t \nabla_{\mathbf{x}_t} \log p_t(\mathbf{x}_t) \mathrm{d}t, \ \ \mathbf{x}_T \sim p_T(\mathbf{x}_T)
    \end{equation}
    Using the posterior mean $\mathbb{E}[\mathbf{x}_0|\mathbf{x}_t]$ to represent the score $\nabla_{\mathbf{x}_t} \log p_t(\mathbf{x}_t)$ using Lemma~\ref{lemma:tweedie}, we recover Eq.~\eqref{eq:uncond samp}:
    \begin{equation}
        \mathrm{d}\mathbf{x}_t = \frac{\mathbf{x}_t - \mathbb{E}[\mathbf{x}_0|\mathbf{x}_t]}{t} \mathrm{d}t, \ \ \mathbf{x}_T \sim p_T(\mathbf{x}_T)
    \end{equation}
    
    Likewise, the following ODE preserves the marginal $p_t(\mathbf{x}_t|\mathbf{y})$ for all $t\in [0, T]$:
    \begin{equation}
        \mathrm{d}\mathbf{x}_t = -t \nabla_{\mathbf{x}_t} \log p_t(\mathbf{x}_t|\mathbf{y}) \mathrm{d}t, \ \ \mathbf{x}_T \sim p_T(\mathbf{x}_T|\mathbf{y})
    \end{equation}
    By Lemma~\ref{lemma:cond-tweedie}, we recover Eq.~\eqref{eq: cond sample}:
    \begin{equation}
        \mathrm{d}\mathbf{x}_t = \frac{\mathbf{x}_t - \mathbb{E}[\mathbf{x}_0|\mathbf{x}_t, \mathbf{y}]}{t} \mathrm{d}t, \ \ \mathbf{x}_T \sim p_T(\mathbf{x}_T|\mathbf{y})
    \end{equation}
\end{proof}

\subsection{Proof of Proposition~\ref{prop:exy}}\label{app:prop.exy}
To relate $\mathbb{E}[\mathbf{x}_0|\mathbf{x}_t, \mathbf{y}]$ to $\mathbb{E}[\mathbf{x}_0|\mathbf{x}_t]$, we note that 
\begin{equation}
    \nabla_{\mathbf{x}_t} \log p_t(\mathbf{x}_t|\mathbf{y}) = \nabla_{\mathbf{x}_t} \log p_t(\mathbf{x}_t) + \nabla_{\mathbf{x}_t} \log p_t(\mathbf{y}|\mathbf{x}_t)
\end{equation}
Using Lemma~\ref{lemma:tweedie} and Lemma~\ref{lemma:cond-tweedie}, we have
\begin{equation}
    \frac{1}{s_t^2 \sigma_t^2}(s_t \mathbb{E}[\mathbf{x}_0|\mathbf{x}_t, \mathbf{y}] - \mathbf{x}_t) = \frac{1}{s_t^2 \sigma_t^2}(s_t \mathbb{E}[\mathbf{x}_0|\mathbf{x}_t] - \mathbf{x}_t) +  \nabla_{\mathbf{x}_t} \log p_t(\mathbf{y}|\mathbf{x}_t)
\end{equation}
and consequently,
\begin{equation}
    \mathbb{E}[\mathbf{x}_0|\mathbf{x}_t, \mathbf{y}] = \mathbb{E}[\mathbf{x}_0|\mathbf{x}_t] + s_t \sigma_t^2 \nabla_{\mathbf{x}_t} \log p_t(\mathbf{y}|\mathbf{x}_t)
\end{equation}

\subsection{Proof of Proposition~\ref{prop:ddnm}}\label{app:proof.ddnm}
When $\sigma \rightarrow 0$, finding the minimizer of Eq.~\eqref{eq:proximal} is equivalent to solving the following hard-constraint optimization problem:
\begin{equation}
    \min_{\hat{\mathbf{x}}_0} \lVert \hat{\mathbf{x}}_0 - D_t(\mathbf{x}_t) \rVert^2_2 \ \ \ \ \mathrm{s.t.} \ \ \ \ \mathbf{y} = \mathbf{A} \hat{\mathbf{x}}_0
\end{equation}
We define the Lagrangian $\mathcal{L}(\hat{\mathbf{x}}_0, \lambda) =  \frac{1}{2}\lVert \hat{\mathbf{x}}_0 - D_t(\mathbf{x}_t) \rVert^2_2 + \lambda^T (\mathbf{y} - \mathbf{A} \hat{\mathbf{x}}_0)$, where $\lambda$ is the Lagrangian multiplier. By the optimality condition, we have
\begin{align}
    & \nabla_{\hat{\mathbf{x}}_0}\mathcal{L} = \hat{\mathbf{x}}_0 - D_t(\mathbf{x}_t) - \mathbf{A}^T \lambda = \mathbf{0} \label{eq: lag1}\\
    & \nabla_{\lambda}\mathcal{L} =  \mathbf{y} - \mathbf{A} \hat{\mathbf{x}}_0 = \mathbf{0} \label{eq: lag2}
\end{align}
Multiplying $\mathbf{A}$ to Eq.~\eqref{eq: lag1} and combining the condition of Eq.~\eqref{eq: lag2} gives:
\begin{align}
    & \mathbf{A}(\hat{\mathbf{x}}_0 - D_t(\mathbf{x}_t) - \mathbf{A}^T \lambda) = \mathbf{0} \\
    & \Rightarrow \mathbf{y} - \mathbf{A}D_t(\mathbf{x}_t) - \mathbf{A}\mathbf{A}^T \lambda = \mathbf{0} \\
    & \Rightarrow \mathbf{A}\mathbf{A}^T\lambda = \mathbf{y} - \mathbf{A}D_t(\mathbf{x}_t) \label{eq:lam}
\end{align}
Multiplying $\mathbf{A}^{\dagger}$ to Eq.~\eqref{eq:lam} and leveraging the property $\mathbf{A}^{\dagger}\mathbf{A}\mathbf{A}^T=\mathbf{A}^T$, we have
\begin{equation}
    \mathbf{A}^T\lambda = \mathbf{A}^{\dagger} \mathbf{y} - \mathbf{A}^{\dagger}\mathbf{A}D_t(\mathbf{x}_t)
\end{equation}
and consequently,
\begin{align}
   \hat{\mathbf{x}}_0 &= D_t(\mathbf{x}_t) + \mathbf{A}^{\dagger} \mathbf{y} - \mathbf{A}^{\dagger}\mathbf{A}D_t(\mathbf{x}_t) \\
    &=  \mathbf{A}^{\dagger}\mathbf{y} + (\mathbf{I} - \mathbf{A}^{\dagger}\mathbf{A})D_t(\mathbf{x}_t)
\end{align}

\subsection{Proof of Theorem~\ref{thm:relation of pos and rev var}}
\label{sec:pf relation of pos and rev var}
We follow~\cite{bao2022analyticdpm} to derive the relationship between the optimal reverse variances $\mathbf{v}_t^2(\mathbf{x}_t)$ and optimal posterior variances $\mathbf{r}_t^2(\mathbf{x}_t)$ based on more general non-Markov forward process introduced by~\cite{song2021denoising}. To find the optimal solution to Eq.~\eqref{eq:vae-elbo0}, we present a much simpler proof than \cite{bao2022analyticdpm} using functional derivatives motivated by \cite{rezende2018taming}. Given a noise schedule $\{\beta_t\}_{t=1}^T$ and $\alpha_t = 1 - \beta_t$, the forward process $q(\mathbf{x}_{1:T}|\mathbf{x}_0)$ is defined as
\begin{align}
    & q(\mathbf{x}_{1:T}|\mathbf{x}_0) = q(\mathbf{x}_T|\mathbf{x}_0)\prod_{t=2}^{T} q(\mathbf{x}_{t-1}|\mathbf{x}_t,\mathbf{x}_0) \\
    & q(\mathbf{x}_{t-1}|\mathbf{x}_t,\mathbf{x}_0) = \mathcal{N}(\mathbf{x}_{t-1}|\Tilde{\mu}_t(\mathbf{x}_t,\mathbf{x}_0), \lambda_t^2 \mathbf{I}) \\
    & \Tilde{\mu}_t(\mathbf{x}_t,\mathbf{x}_0) = \sqrt{\bar{\alpha}_{t-1}}\mathbf{x}_0 + \sqrt{\bar{\beta}_{t-1} - \lambda_t^2}\cdot\frac{\mathbf{x}_t - \sqrt{\bar{\alpha}_t}\mathbf{x}_0}{\sqrt{\bar{\beta}_t}} 
\end{align}
where $\bar{\alpha}_t = \prod_{i=1}^{t} \alpha_i$ and $\bar{\beta}_t = 1 - \bar{\alpha}_t$. \cite{song2021denoising} show that for arbitrary choice of $\lambda_t$, the marginal distributions $q(\mathbf{x}_t|\mathbf{x}_0)$ maintain $q(\mathbf{x}_t|\mathbf{x}_0)=\mathcal{N}(\mathbf{x}_t|\sqrt{\bar{\alpha}_t}\mathbf{x}_0, \bar{\beta}_t \mathbf{I})$. DDPM forward process is a special case when $\lambda_t^2 = \Tilde{\beta}_t$ with $\Tilde{\beta}_t = \frac{\bar{\beta}_{t-1}}{\bar{\beta}_t}\beta_t$, which is used in~\cite{nichol2021improved} for pre-training DDPM model.

To fit the data distribution $q(\mathbf{x}_0)$, we define the reverse process, given by
\begin{equation}
    p(\mathbf{x}_{0:T}) = p(\mathbf{x}_T)\prod_{t=1}^T p(\mathbf{x}_{t-1}|\mathbf{x}_t), \ \ p(\mathbf{x}_{t-1}|\mathbf{x}_t) = \mathcal{N}(\mathbf{x}_{t-1}|\mathbf{m}_t(\mathbf{x}_t),\mathrm{diag}[\mathbf{v}_t^2(\mathbf{x}_t)])
\end{equation}

To train $p(\mathbf{x}_0)$ we minimize the KL divergence between the forward and the reverse process:
\begin{equation}
\label{eq:vae-elbo}
    \min_p D_{KL}(q(\mathbf{x}_{0:T}) || p(\mathbf{x}_{0:T})), \ \ q(\mathbf{x}_{0:T}) = q(\mathbf{x}_0) q(\mathbf{x}_{1:T}|\mathbf{x}_0)
\end{equation}
which is equivalent to minimizing the variational bound $\mathbb{E}_q[L_{\text{vb}}]$ on negative log-likelihood of data distribution $q(\mathbf{x}_0)$ with $L_{\text{vb}}$ given as follows:
\begin{align}
    & L_{\text{vb}} = L_0 + L_1 + ... + L_T \\
    & L_0 = -\log p(\mathbf{x}_0|\mathbf{x}_1) \\
    & L_{t-1} = D_{KL}(q(\mathbf{x}_{t-1}|\mathbf{x}_0, \mathbf{x}_t)||p(\mathbf{x}_{t-1}|\mathbf{x}_t)) \\
    & L_T =  D_{KL}(q(\mathbf{x}_{T}|\mathbf{x}_0)||p(\mathbf{x}_T))
\end{align}

For $t\in [2, T]$, $L_{t-1}$ are KL divergences between two Gaussians, which possess analytical forms:
\begin{align}
    L_{t-1} &\equiv \log \frac{|\mathrm{diag}[\mathbf{v}_t^2(\mathbf{x}_t)]|}{|\lambda_t^2 \mathbf{I}|} + \lVert \Tilde{\mu}_t(\mathbf{x}_t,\mathbf{x}_0) - \mathbf{m}(\mathbf{x}_t)\rVert^2_{\mathrm{diag}[\mathbf{v}_t^2(\mathbf{x}_t)]^{-1}} + \mathrm{tr}[\lambda_t^2 \mathrm{diag}[\mathbf{v}_t^2(\mathbf{x}_t)]^{-1}] \\
    &\equiv \sum_{i=1}^d \log \mathbf{v}_t^2(\mathbf{x}_t)_i + \frac{(\Tilde{\mu}_t(\mathbf{x}_t,\mathbf{x}_0)_i - \mathbf{m}_t(\mathbf{x}_t)_i)^2}{\mathbf{v}_t^2(\mathbf{x}_t)_i} + \frac{\lambda_t^2}{\mathbf{v}_t^2(\mathbf{x}_t)_i} 
\end{align}
where ``$\equiv$'' denotes ``equals up to a constant and a scaling factor'' and $i$ indexes the elements of an vector. 

Note that minimizing $\mathbb{E}_q[L_{\text{vb}}]$ can be decomposed into $T$ independent optimization sub-problems: 
\begin{equation}
    \min_{\mathbf{m}_t, \mathbf{v}_t} \mathbb{E}_{q(\mathbf{x}_0, \mathbf{x}_t)}[L_{t-1}], \ \ t \in [1, T]
\end{equation}

The optimal $\mathbf{m}_t$ and $\mathbf{v}_t$ can be found by taking the functional derivatives of $\mathbb{E}_{q(\mathbf{x}_0, \mathbf{x}_t)}[L_{t-1}]$ w.r.t $\mathbf{m}_t$ and $\mathbf{v}_t^2$ then set to zero:
\begin{align}
    &\frac{\delta \mathbb{E}_{q(\mathbf{x}_0, \mathbf{x}_t)}[L_{t-1}]}{\delta \mathbf{m}_t(\mathbf{x}_t)_i} \equiv \mathbb{E}_{q(\mathbf{x}_0)} [q(\mathbf{x}_t|\mathbf{x}_0) \frac{\mathbf{m}_t(\mathbf{x}_t)_i - \Tilde{\mu}_t(\mathbf{x}_t,\mathbf{x}_0)_i}{\mathbf{v}_t^2(\mathbf{x}_t)_i}] = 0  \label{eq:func deriv v 1}\\
     &\frac{\delta \mathbb{E}_{q(\mathbf{x}_0, \mathbf{x}_t)}[L_{t-1}]}{\delta \mathbf{v}_t^2(\mathbf{x}_t)_i} \equiv \mathbb{E}_{q(\mathbf{x}_0)} [q(\mathbf{x}_t|\mathbf{x}_0) (\frac{1}{\mathbf{v}_t^2(\mathbf{x}_t)_i} - \frac{(\Tilde{\mu}_t(\mathbf{x}_t,\mathbf{x}_0)_i - \mathbf{m}(\mathbf{x}_t)_i)^2}{(\mathbf{v}_t^2(\mathbf{x}_t)_i)^2} - \frac{\lambda_t^2}{(\mathbf{v}_t^2(\mathbf{x}_t)_i)^2})] = 0 \label{eq:func deriv v 2}
\end{align}

We can solve for optimal $\mathbf{m}_t(\mathbf{x}_t)_i$ by rearranging Eq.~\eqref{eq:func deriv v 1}:
\begin{align}
    &\mathbb{E}_{q(\mathbf{x}_0)}[q(\mathbf{x}_t|\mathbf{x}_0)] \mathbf{m}_t(\mathbf{x}_t)_i = \mathbb{E}_{q(\mathbf{x}_0)}[q(\mathbf{x}_t|\mathbf{x}_0) \Tilde{\mu}_t(\mathbf{x}_t,\mathbf{x}_0)_i] \\
    &\Rightarrow q(\mathbf{x}_t) \mathbf{m}_t(\mathbf{x}_t)_i = \int q(\mathbf{x}_0)q(\mathbf{x}_t|\mathbf{x}_0) \Tilde{\mu}_t(\mathbf{x}_t,\mathbf{x}_0)_i \mathrm{d}\mathbf{x}_0 \\
    &\Rightarrow \mathbf{m}_t(\mathbf{x}_t)_i = \int q(\mathbf{x}_0|\mathbf{x}_t) \Tilde{\mu}_t(\mathbf{x}_t,\mathbf{x}_0)_i \mathrm{d}\mathbf{x}_0 \\
    &\Rightarrow \mathbf{m}_t(\mathbf{x}_t)_i = \Tilde{\mu}_t(\mathbf{x}_t, \mathbb{E}[\mathbf{x}_0|\mathbf{x}_t])_i \label{eq:opt m} 
\end{align}
where Eq.~\eqref{eq:opt m} is due to the linearity of $\Tilde{\mu}_t$ w.r.t $\mathbf{x}_0$.

Likewise, rearranging Eq.~\eqref{eq:func deriv v 2} gives
\begin{equation}
    \mathbf{v}_t^2(\mathbf{x}_t)_i = \lambda_t^2 + \mathbb{E}_{q(\mathbf{x}_0|\mathbf{x}_t)}[(\Tilde{\mu}_t(\mathbf{x}_t,\mathbf{x}_0)_i - \mathbf{m}(\mathbf{x}_t)_i)^2] \label{eq:opt v}
\end{equation}

By plugging the optimal $\mathbf{m}_t(\mathbf{x}_t)_i$ determined by Eq.~\eqref{eq:opt m} into Eq.~\eqref{eq:opt v} and dropping the element index $i$, we conclude the proof:
\begin{align}
    \mathbf{v}_t^2(\mathbf{x}_t) &= \lambda_t^2 + \mathbb{E}_{q(\mathbf{x}_0|\mathbf{x}_t)}[(\Tilde{\mu}_t(\mathbf{x}_t,\mathbf{x}_0) - \Tilde{\mu}_t(\mathbf{x}_t, \mathbb{E}[\mathbf{x}_0|\mathbf{x}_t]))^2] \\
    &= \lambda_t^2 + \mathbb{E}_{q(\mathbf{x}_0|\mathbf{x}_t)}[(\Tilde{\mu}_t(\mathbf{0},\mathbf{x}_0 - \mathbb{E}[\mathbf{x}_0|\mathbf{x}_t]))^2] \\
    &= \lambda_t^2 + (\sqrt{\bar{\alpha}_{t-1}} - \sqrt{\bar{\beta}_{t-1} - \lambda_t^2} \sqrt{\frac{\bar{\alpha}_t}{\bar{\beta}_t}})^2 \cdot \mathbb{E}_{q(\mathbf{x}_0|\mathbf{x}_t)}[(\mathbf{x}_0 - \mathbb{E}[\mathbf{x}_0|\mathbf{x}_t])^2] \\
    &= \lambda_t^2 + (\sqrt{\bar{\alpha}_{t-1}} - \sqrt{\bar{\beta}_{t-1} - \lambda_t^2} \sqrt{\frac{\bar{\alpha}_t}{\bar{\beta}_t}})^2 \cdot \mathbf{r}_t^2(\mathbf{x}_t)
\end{align}

We are often given a pre-trained DDPM model with learned reverse variances. The following Corollary of Theorem~\ref{thm:relation of pos and rev var} gives a simplified relationship between optimal posterior variances and optimal reverse variances under DDPM case:
\begin{corollary}
For the DDPM forward process $\lambda_t^2 = \Tilde{\beta}_t$, the optimal posterior variances $\mathbf{r}_t^{*2}(\mathbf{x}_t)$ and optimal reverse variances $\mathbf{v}_t^{*2}(\mathbf{x}_t)$ are related by
\begin{equation}
    \mathbf{v}_t^{*2}(\mathbf{x}_t) = \Tilde{\beta}_t + (\frac{\sqrt{\bar{\alpha}_{t-1}}\beta_t}{1-\bar{\alpha}_t})^{2} \cdot \mathbf{r}_t^{*2}(\mathbf{x}_t)
\end{equation}
\end{corollary}

\begin{remark}
According Theorem~\ref{thm:relation of pos and rev var}, to perform optimal ancestral sampling, we only need to provide the MMSE estimator $\mathbb{E}[\mathbf{x}_0|\mathbf{x}_t]$ to compute the reverse mean $\mathbf{m}^{*}_t(\mathbf{x}_t)$ and the posterior variances $\mathbf{r}^{*2}_t(\mathbf{x}_t)$ to compute the reverse variances $\mathbf{v}^{*2}_t(\mathbf{x}_t)$, which can be both obtained from the proposed pre-training~(Eq.~\eqref{eq:klobj}). This may provide an alternative way for pre-trainning DDPM model that differs from~\cite{nichol2021improved}.
\end{remark}

\subsection{Proof of Proposition~\ref{prop:opt-klobj}}\label{app:proof.opt-klobj}
    Deriving the optimal solution to~\text{Eq.~\eqref{eq:klobj}} under the diagonal posterior covariance case, i.e., $\Sigma_t(\mathbf{x}_t)=\mathrm{diag}[\mathbf{r}_t^2(\mathbf{x}_t)]$, is similar to Appendix~\ref{sec:pf relation of pos and rev var}. Note that we seek for point-wise minimizer of Eq.~\eqref{eq:klobj}, i.e., find the optimum of
    \begin{equation}\label{eq:diag-obj}
        \mathbb{E}_{p_t(\mathbf{x}_0, \mathbf{x}_t)}[\log q_t(\mathbf{x}_0|\mathbf{x}_t)] \equiv \mathbb{E}_{p_t(\mathbf{x}_0, \mathbf{x}_t)}\left[\sum_{i=1}^d \frac{1}{\mathbf{r}_t^2(\mathbf{x}_t)_i}(\mathbf{x}_{0}-D_t(\mathbf{x}_t))_i^2 + \log \mathbf{r}_t^2(\mathbf{x}_t)_i\right]
    \end{equation}
    
    For any $t$, taking the functional derivatives of $\mathbb{E}_{p_t(\mathbf{x}_0, \mathbf{x}_t)}[\log q_t(\mathbf{x}_0|\mathbf{x}_t)]$ w.r.t $D_t(\mathbf{x}_t)_i$ and $\mathbf{r}_t^2(\mathbf{x}_t)_i$ and then set to zero, we obtain the optimality conditions:
    \begin{align}
        &\mathbb{E}_{p(\mathbf{x}_0)}[p_t(\mathbf{x}_t|\mathbf{x}_0)(D_t(\mathbf{x}_t)_i - \mathbf{x}_{0i})] = 0 \label{eq:mle-opt-cond-1}\\
        &\mathbb{E}_{p(\mathbf{x}_0)}[p_t(\mathbf{x}_t|\mathbf{x}_0)(-\frac{1}{(\mathbf{r}_t^2(\mathbf{x}_t)_i)^2}(\mathbf{x}_{0}-D_t(\mathbf{x}_t))_i^2 + \frac{1}{\mathbf{r}_t^2(\mathbf{x}_t)_i}] = 0 \label{eq:mle-opt-cond-2}
    \end{align}
    Combining the optimality conditions given by Eq.~\eqref{eq:mle-opt-cond-1} and Eq.~\eqref{eq:mle-opt-cond-2}, we conclude the proof.

\section{Practical Numerical Algorithms}\label{sec:numerical}
In this section, we discuss practical algorithms for implementing efficient Type I and Type II guidance.

\textbf{Type I guidance.} We are required to obtain the likelihood score, which can be approximated via Jacobian-vector product (JVP) similar to~\cite{song2023pseudoinverseguided}:
\begin{equation}\label{eq:vjp}
    \nabla_{\mathbf{x}_t} \log p(\mathbf{y}|\mathbf{x}_t) \approx \frac{\partial D_t(\mathbf{x}_t)}{\partial \mathbf{x}_t} \underbrace{\mathbf{A}^T(\sigma^2\mathbf{I} + \mathbf{A} \Sigma_t(\mathbf{x}_t) \mathbf{A}^T)^{-1}(\mathbf{y}- \mathbf{A} D_t(\mathbf{x}_t))}_{\mathbf{v}}.
\end{equation} 

\textbf{Type II guidance.} We are required to solve the following \textit{auto-weighted} proximal problem: 
\begin{equation}
\label{eq:impt2-app}
    \mathbf{x}_0^{(t)} = \arg\min_{\mathbf{x}_0} \lVert \mathbf{y} -\mathbf{A}\mathbf{x}_0 \rVert^2 + \sigma^2 \lVert \mathbf{x}_0 - D_t(\mathbf{x}_t) \rVert^2_{\Sigma_t^{-1}(\mathbf{x}_t)}
\end{equation}
which has general closed-form solution given by
\begin{equation}
\label{eq:sol impt2}
    \mathbf{x}_0^{(t)} = (\Sigma_t(\mathbf{x}_t)^{-1} + \frac{1}{\sigma^2}\mathbf{A}^T \mathbf{A})^{-1}( \Sigma_t(\mathbf{x}_t)^{-1}D_t(\mathbf{x}_t) + \frac{1}{\sigma^2}\mathbf{A}^T \mathbf{y}).
\end{equation}
The derivation of Eq.~\eqref{eq:sol impt2} can be directly obtained by computing the mean of $q_t(\mathbf{x}_0|\mathbf{x}_t, \mathbf{y})$ using the Bayes' theorem for Gaussian variables~\cite{bishop2006pattern}. Leveraging the Woodbury matrix identity, we have an equivalent form\footnote{Note that this form also share similar mathematical form to noisy version of DDNM, i.e., DDNM$^+$~(Eq.~(17), \citet{wang2023zeroshot}).} for Eq.~\eqref{eq:sol impt2}, as follows:
\begin{equation} \label{eq:wood-2}
    \mathbf{x}_0^{(t)} = D_t(\mathbf{x}_t)+ \Sigma_t(\mathbf{x}_t) \underbrace{\mathbf{A}^T (\sigma^2 \mathbf{I} + \mathbf{A} \Sigma_t(\mathbf{x}_t) \mathbf{A}^T)^{-1} (\mathbf{y} - \mathbf{A}D_t(\mathbf{x}_t))}_{\mathbf{v}}.
\end{equation}

To evaluate Eq.~\eqref{eq:vjp} and Eq.~\eqref{eq:wood-2} for realizing Type I and Type II guidance, the computational challenge arises from evaluating the high-dimensional matrix inversion for $\sigma^2 \mathbf{I} + \mathbf{A} \Sigma_t(\mathbf{x}_t) \mathbf{A}^T$, which in general is $\mathcal{O}(d^3)$. To address this, note that Eq.~\eqref{eq:vjp} and Eq.~\eqref{eq:wood-2} can be efficiently evaluated if we have fast matrix-vector multiplies available for $\mathbf{v} = \mathbf{A}^T(\sigma^2 \mathbf{I} + \mathbf{A} \Sigma_t(\mathbf{x}_t) \mathbf{A}^T)^{-1} (\mathbf{y} - \mathbf{A}D_t(\mathbf{x}_t))$. In the subsequent sections, we will discuss two methods based on closed-form solutions and CG for fast computation for $\mathbf{v}$, respectively designed for isotropic covariances and more general covariances.

\vspace{10pt}

\begin{remark}
The discussions in this section indeed reveal a deeper connection in existing approaches. Suppose $D_t(\mathbf{x}_t)= \mathbb{E}[\mathbf{x}_0|\mathbf{x}_t]$. Then, by Proposition~\ref{prop:exy}, Eq.~\eqref{eq:vjp}, and the second-order Tweedie's formula $\sigma_t^2 \frac{\partial \mathbb{E}[\mathbf{x}_0|\mathbf{x}_t]}{\partial \mathbf{x}_t} = \Sigma_t^*(\mathbf{x}_t)$, the conditional posterior mean approximation in Type I guidance can be rewritten as:
\begin{equation} \label{eq:t1-uni}
     \mathbf{x}_0^{(t)} = D_t(\mathbf{x}_t) + \Sigma_t^*(\mathbf{x}_t) \mathbf{A}^T (\sigma^2 \mathbf{I} + \mathbf{A} \Sigma_t(\mathbf{x}_t) \mathbf{A}^T)^{-1} (\mathbf{y} - \mathbf{A}D_t(\mathbf{x}_t)).
\end{equation}
\end{remark}
As can be seen, Eq.~\eqref{eq:t1-uni} of Type I and Eq.~\eqref{eq:wood-2} of Type II can be uniformly considered as an approximation for the mean of the optimal approximated conditional denoising posterior $q_t(\mathbf{x}_0|\mathbf{x}_t, \mathbf{y})\propto p(\mathbf{y}|\mathbf{x}_0)\mathcal{N}(\mathbf{x}_0|\mathbb{E}[\mathbf{x}_0|\mathbf{x}_t], \Sigma_t^*(\mathbf{x}_t))$, i.e.,
\begin{equation}
     \mathbb{E}[\mathbf{x}_0|\mathbf{x}_t] + \Sigma_t^*(\mathbf{x}_t) \mathbf{A}^T (\sigma^2 \mathbf{I} + \mathbf{A} \Sigma_t^*(\mathbf{x}_t) \mathbf{A}^T)^{-1} (\mathbf{y} - \mathbf{A}\mathbb{E}[\mathbf{x}_0|\mathbf{x}_t]).
\end{equation}

\subsection{Using Closed-form Solutions for Isotropic Posterior Covariance}
\label{sec:solu-g}
In this section, we provide efficient closed-form results for computing $\mathbf{v}$ under isotropic posterior covariance: $\Sigma_t(\mathbf{x}_t)=r_t^2(\mathbf{x}_t) \mathbf{I}$. Before we delve into the closed-form results, we first give some important notations. We define the downsampling operator given sampling position $\mathbf{m} \in \{0, 1\}^{d\times 1}$ as $\mathbf{D}_{\mathbf{m}} \in \{0, 1\}^{\lVert m\rVert_0 \times d}$, which selects rows of a given matrix that correspond to one in $\mathbf{m}$ and when performing left multiplication. We use $\mathbf{D}_{\downarrow s}$ to denote the standard $s$-folds downsampling operator, which is equivalent to $\mathbf{D}_{\mathbf{m}}$ when ones in $\mathbf{m}$ are spaced evenly. For image signal, it selects the upper-left pixel for each distinct $s\times s$ patch~\cite{zhang2020deep} when performing left multiplication to the \textit{vectorized} image. We use $\mathbf{D}_{\Downarrow s}$ to denote the distinct block downsampler, i.e., averaging $s$ length $d/s$ distinct blocks of a vector. For image signal, it averaging distinct $d/s\times d/s$ blocks~\cite{zhang2020deep} when performing left multiplication to the \textit{vectorized} image. We denote the Fourier transform matrix for $d$-dimensional signal as $\mathbf{F}$, Fourier transform matrix for $d/s$-dimensional signal as $\mathbf{F}_{\downarrow s}$, the Fourier transform of a vector $\mathbf{v}$ as $\hat{\mathbf{v}}$, and the complex conjugate of a complex vector $\mathbf{v}$ as $\bar{\mathbf{v}}$. We use $\odot$ to denote element-wise multiplication, and the divisions used below are also element-wise.

\begin{lemma}
\label{lemma:FDF}
    Performing $s$-fold standard downsampling in spacial domain is equivalent to performing $s$-fold block downsampling in frequency domain: $\mathbf{D}_{\Downarrow s}=\mathbf{F}_{\downarrow s}\mathbf{D}_{\downarrow s} \mathbf{F}^{-1}$.
    \begin{proof}
    Considering an arbitrary $d$-dimensional signal in frequency domain $\hat{\mathbf{x}}[k], \ k=0,2,..,d-1$. Multiplying $\mathbf{F}_{\downarrow s}\mathbf{D}_{\downarrow s} \mathbf{F}^{-1}$ to $\hat{\mathbf{x}}$ is equivalent to letting $\hat{\mathbf{x}}$ go through the following linear system and obtain the output $\hat{\mathbf{y}}$:
    \begin{align}
        &\mathbf{x}[n] = \frac{1}{d}\sum_{k=0}^{d-1} \hat{\mathbf{x}}[k] e^{j\frac{2\pi}{d}kn} \\
        &\mathbf{x}_{\downarrow s}[n] = \mathbf{x}[ns] \\
        &\hat{\mathbf{y}}[k] = \sum_{n=0}^{d/s - 1} \mathbf{x}_{\downarrow s}[n] e^{-j\frac{2\pi}{d/s}kn}
    \end{align}
    We now use $\hat{\mathbf{x}}$ to represent $\hat{\mathbf{y}}$:
    \begin{align}
        \hat{\mathbf{y}}[k] &= \sum_{n=0}^{d/s - 1} \frac{1}{d}\sum_{k'=0}^{d-1} \hat{\mathbf{x}}[k'] e^{j\frac{2\pi}{d}k'ns} e^{-j\frac{2\pi}{d/s}kn} \\
        &= \sum_{n=0}^{d/s - 1} \frac{1}{d} e^{-j\frac{2\pi}{d/s}kn} \sum_{k'=0}^{d-1} \hat{\mathbf{x}}[k'] e^{j\frac{2\pi}{d/s}k'n} \\
        &= \sum_{n=0}^{d/s - 1} \frac{1}{d} e^{-j\frac{2\pi}{d/s}kn} (\sum_{k'=0}^{d/s-1} + \sum_{k'=d/s}^{2d/s - 1} + \sum_{k'=2d/s}^{3d/s - 1} + ... + \sum_{k'=(s-1)d/s}^{sd/s-1}) \hat{\mathbf{x}}[k'] e^{j\frac{2\pi}{d/s}k'n} \\
        &= \sum_{n=0}^{d/s - 1} \frac{1}{d} e^{-j\frac{2\pi}{d/s}kn} \sum_{k'=0}^{d/s-1} (\hat{\mathbf{x}}[k'] + \hat{\mathbf{x}}[k'+d/s] + ... + \hat{\mathbf{x}}[k'+(s-1)d/s]) e^{j\frac{2\pi}{d/s}k'n} \label{eq:down-in-space-1} \\
        &=\sum_{n=0}^{d/s - 1} e^{-j\frac{2\pi}{d/s}kn} \frac{1}{d/s} \sum_{k'=0}^{d/s-1} \frac{\hat{\mathbf{x}}[k'] + \hat{\mathbf{x}}[k'+d/s] + ... + \hat{\mathbf{x}}[k'+(s-1)d/s]}{s} e^{j\frac{2\pi}{d/s}k'n} \\
        &= \frac{\hat{\mathbf{x}}[k] + \hat{\mathbf{x}}[k+d/s] + ... + \hat{\mathbf{x}}[k+(s-1)d/s]}{s} \label{eq:down-in-space-2}
    \end{align}
    where Eq.~\eqref{eq:down-in-space-1} is because $e^{j\frac{2\pi}{d/s}k'n}$ has period of $d/s$ in $k'$ and Eq.~\eqref{eq:down-in-space-2} is because $d/s$-dimensional inverse Fourier transform and Fourier transform are canceled out. From Eq.~\eqref{eq:down-in-space-2} we have $\hat{\mathbf{y}} = \mathbf{D}_{\Downarrow s}\hat{\mathbf{x}}$. So $\mathbf{D}_{\Downarrow s}\hat{\mathbf{x}}=\mathbf{F}_{\downarrow s}\mathbf{D}_{\downarrow s} \mathbf{F}^{-1}\hat{\mathbf{x}}$ for any $\hat{\mathbf{x}}$, and consequently, $\mathbf{D}_{\Downarrow s}=\mathbf{F}_{\downarrow s}\mathbf{D}_{\downarrow s} \mathbf{F}^{-1}$.
    \end{proof}
\end{lemma}

\textbf{Inpainting.} The observation model for image inpainting can be expressed as:
\begin{equation}
\label{eq:model-inpaint}
    \mathbf{y} = \underbrace{\mathbf{D}_{\mathbf{m}}}_{\mathbf{A}} \mathbf{x}_0 + \mathbf{n},
\end{equation}
and the closed-form solution to $\mathbf{v}$ in image inpainting is given by the following:
\begin{equation}
    \mathbf{v} = \frac{\Tilde{\mathbf{y}} - \mathbf{m} \odot D_t(\mathbf{x}_t)}{\sigma^2 + r_t^2(\mathbf{x}_t)}
\end{equation}
where $\Tilde{\mathbf{y}} = \mathbf{D}_{\mathbf{m}}^T \mathbf{y} = \mathbf{m} \odot (\mathbf{x}_0 + \Tilde{\mathbf{n}}), \, \Tilde{\mathbf{n}} \sim \mathcal{N}(\mathbf{0}, \mathbf{I})$ is the zero-filling measurements that fills the masked region with zeros and processes the exact same size to $\mathbf{x}_0$. In practice, the measurements in inpainting are usually stored in the form of $\Tilde{\mathbf{y}}$, while $\mathbf{y}$ used here is for mathematical convenience. 
\begin{proof}
    \begin{align}
        \mathbf{v} &= \mathbf{D}_\mathbf{m}^T(\sigma^2 \mathbf{I} + \mathbf{D}_\mathbf{m} r_t^2(\mathbf{x}_t)\mathbf{I} \mathbf{D}_\mathbf{m}^T)^{-1}(\mathbf{y}- \mathbf{D}_\mathbf{m} D_t(\mathbf{x}_t)) \\
        &= \mathbf{D}_\mathbf{m}^T(\sigma^2 \mathbf{I} + r_t^2(\mathbf{x}_t)\mathbf{D}_\mathbf{m} \mathbf{D}_\mathbf{m}^T)^{-1}(\mathbf{y}- \mathbf{D}_\mathbf{m} D_t(\mathbf{x}_t))  \\
        &= \mathbf{D}_\mathbf{m}^T((\sigma^2 + r_t^2(\mathbf{x}_t))\mathbf{I})^{-1}(\mathbf{y}- \mathbf{D}_\mathbf{m} D_t(\mathbf{x}_t)) \label{eq:I-inpaint-2} \\
        &= \frac{\mathbf{D}_\mathbf{m}^T(\mathbf{y}- \mathbf{D}_\mathbf{m}D_t(\mathbf{x}_t))}{\sigma^2 + r_t^2(\mathbf{x}_t)}  \\
        &= \frac{\Tilde{\mathbf{y}}- \mathbf{m}\odot D_t(\mathbf{x}_t)}{\sigma^2 + r_t^2(\mathbf{x}_t)} \label{eq:I-inpaint-4}
    \end{align}
where Eq.~\eqref{eq:I-inpaint-2} and Eq.~\eqref{eq:I-inpaint-4} are because $\mathbf{D}_\mathbf{m} \mathbf{D}_\mathbf{m}^T = \mathbf{I}$ and $\mathbf{D}_\mathbf{m}^T \mathbf{D}_\mathbf{m} = \mathrm{diag}[\mathbf{m}]$.
\end{proof}

\textbf{Debluring.} The observation model for image debluring can be expressed as:
\begin{equation}
\label{eq:deblur-orig}
    \mathbf{y} = \mathbf{x}_0 \ast\mathbf{k} + \mathbf{n},
\end{equation}
where $\mathbf{k}$ is the blurring kernel and $\ast$ is convolution operator. By assuming $\ast$ is a circular convolution operator, we can convert Eq.~\eqref{eq:deblur-orig} to the canonical form $\mathbf{y} = \mathbf{A} \mathbf{x}_0 + \mathbf{n}$ by leveraging the convolution property of Fourier transform:
\begin{equation}
\label{eq:deblur-cano}
    \mathbf{y} = \underbrace{\mathbf{F}^{-1}\mathrm{diag}[\hat{\mathbf{k}}]\mathbf{F}}_{\mathbf{A}}\mathbf{x}_0 + \mathbf{n},
\end{equation}
and the closed-form solution to $\mathbf{v}$ in image debluring is given by the following:
\begin{equation}
    \mathbf{v} = \mathbf{F}^{-1}(\bar{\hat{\mathbf{k}}}\odot\frac{\mathbf{F} (\mathbf{y} - \mathbf{A} D_t(\mathbf{x}_t))}{\sigma^2 + r_t^2(\mathbf{x}_t)\bar{\hat{\mathbf{k}}} \odot \hat{\mathbf{k}}}).
\end{equation}
\begin{proof}
    Since $\mathbf{A}$ is a real matrix, we have $\mathbf{A}^T = \mathbf{A}^H = \mathbf{F}^{-1}\mathrm{diag}[\bar{\hat{\mathbf{k}}}]\mathbf{F}$.
    \begin{align}
        \mathbf{v} &= \mathbf{F}^{-1}\mathrm{diag}[\bar{\hat{\mathbf{k}}}]\mathbf{F} (\sigma^2\mathbf{I} + \mathbf{F}^{-1}\mathrm{diag}[\hat{\mathbf{k}}]\mathbf{F} r_t^2(\mathbf{x}_t)\mathbf{I} \mathbf{F}^{-1}\mathrm{diag}[\bar{\hat{\mathbf{k}}}]\mathbf{F})^{-1}(\mathbf{y} - \mathbf{A} D_t(\mathbf{x}_t)) \\
        &= \mathbf{F}^{-1}\mathrm{diag}[\bar{\hat{\mathbf{k}}}]\mathbf{F} (\sigma^2\mathbf{I} + r_t^2(\mathbf{x}_t)\mathbf{F}^{-1}\mathrm{diag}[\hat{\mathbf{k}}]\mathrm{diag}[\bar{\hat{\mathbf{k}}}]\mathbf{F})^{-1}(\mathbf{y} - \mathbf{A} D_t(\mathbf{x}_t)) \\
        &= \mathbf{F}^{-1}\mathrm{diag}[\bar{\hat{\mathbf{k}}}]\mathbf{F} (\sigma^2\mathbf{I} + r_t^2(\mathbf{x}_t)\mathbf{F}^{-1}\mathrm{diag}[\hat{\mathbf{k}}\odot \bar{\hat{\mathbf{k}}}]\mathbf{F})^{-1}(\mathbf{y} - \mathbf{A} D_t(\mathbf{x}_t)) \\
        &=\mathbf{F}^{-1}\mathrm{diag}[\bar{\hat{\mathbf{k}}}]\mathbf{F} (\mathbf{F}^{-1}(\sigma^2\mathbf{I} + r_t^2(\mathbf{x}_t)\mathrm{diag}[\hat{\mathbf{k}}\odot \bar{\hat{\mathbf{k}}}])\mathbf{F})^{-1}(\mathbf{y} - \mathbf{A} D_t(\mathbf{x}_t)) \\
        &=\mathbf{F}^{-1}\mathrm{diag}[\bar{\hat{\mathbf{k}}}]\mathbf{F} (\mathbf{F}^{-1}\mathrm{diag}[\sigma^2 + r_t^2(\mathbf{x}_t)\hat{\mathbf{k}}\odot \bar{\hat{\mathbf{k}}}]\mathbf{F})^{-1}(\mathbf{y} - \mathbf{A} D_t(\mathbf{x}_t)) \\
        &=\mathbf{F}^{-1}\mathrm{diag}[\bar{\hat{\mathbf{k}}}]\mathbf{F} \mathbf{F}^{-1}\mathrm{diag}[\sigma^2 + r_t^2(\mathbf{x}_t)\hat{\mathbf{k}}\odot \bar{\hat{\mathbf{k}}}]^{-1}\mathbf{F}(\mathbf{y} - \mathbf{A} D_t(\mathbf{x}_t)) \\
        &=\mathbf{F}^{-1}(\bar{\hat{\mathbf{k}}}\odot\frac{\mathbf{F}(\mathbf{y} - \mathbf{A} D_t(\mathbf{x}_t))}{\sigma^2 + r_t^2(\mathbf{x}_t)\hat{\mathbf{k}}\odot \bar{\hat{\mathbf{k}}}})
    \end{align}
\end{proof}

\textbf{Super resolution.} According to \cite{zhang2020deep}, the observation model for image super resolution can be \textit{approximately} expressed as:
\begin{equation}
\label{eq:sr-orig}
    \mathbf{y} = (\mathbf{x}_0 \ast\mathbf{k})_{\downarrow s} + \mathbf{n}
\end{equation}
By leveraging the convolution property of Fourier transform, we can convert Eq.~\eqref{eq:sr-orig} to the canonical form $\mathbf{y} = \mathbf{A} \mathbf{x}_0 + \mathbf{n}$:
\begin{equation}
\label{eq:sr-cano}
    \mathbf{y} = \underbrace{\mathbf{D}_{\downarrow s}\mathbf{F}^{-1}\mathrm{diag}[\hat{\mathbf{k}}]\mathbf{F}}_{\mathbf{A}}\mathbf{x}_0 + \mathbf{n}
\end{equation}
and the closed-form solution to $\mathbf{v}$ in image super resolution is given by the following:
\begin{equation}
    \mathbf{v} = \mathbf{F}^{-1}(\bar{\hat{\mathbf{k}}} \odot_s \frac{\mathbf{F}_{\downarrow_s} (\mathbf{y} - \mathbf{A} D_t(\mathbf{x}_t))}{\sigma^2 + r_t^2(\mathbf{x}_t)( \bar{\hat{\mathbf{k}}} \odot \hat{\mathbf{k}})_{\Downarrow_s}})
\end{equation}
where $\odot_s$ denotes block processing operator with element-wise multiplication~\cite{zhang2020deep}.

\begin{proof}
    Since $\mathbf{A}$ is a real matrix, we have $\mathbf{A}^T = \mathbf{A}^H = \mathbf{F}^{-1}\mathrm{diag}[\bar{\hat{\mathbf{k}}}]\mathbf{F} \mathbf{D}_{\downarrow s}^T$, then
    \begin{align}
        \mathbf{v} &= \mathbf{F}^{-1}\mathrm{diag}[\bar{\hat{\mathbf{k}}}]\mathbf{F} \mathbf{D}_{\downarrow s}^T(\sigma^2 \mathbf{I} + \mathbf{D}_{\downarrow s}\mathbf{F}^{-1}\mathrm{diag}[\hat{\mathbf{k}}]\mathbf{F} r_t^2(\mathbf{x}_t)\mathbf{I} \mathbf{F}^{-1}\mathrm{diag}[\bar{\hat{\mathbf{k}}}]\mathbf{F} \mathbf{D}_{\downarrow s}^T)^{-1}(\mathbf{y}-\mathbf{A}D_t(\mathbf{x}_t)) \\
        &= \mathbf{F}^{-1}\mathrm{diag}[\bar{\hat{\mathbf{k}}}]\mathbf{F} \mathbf{D}_{\downarrow s}^T(\sigma^2 \mathbf{I} + r_t^2(\mathbf{x}_t)\mathbf{D}_{\downarrow s}\mathbf{F}^{-1}\mathrm{diag}[\hat{\mathbf{k}}\odot \bar{\hat{\mathbf{k}}}]\mathbf{F} \mathbf{D}_{\downarrow s}^T)^{-1}(\mathbf{y}-\mathbf{A}D_t(\mathbf{x}_t)) \\
        &= \mathbf{F}^{-1}\mathrm{diag}[\bar{\hat{\mathbf{k}}}]\mathbf{F} \mathbf{D}_{\downarrow s}^T(\mathbf{D}_{\downarrow s}\mathbf{F}^{-1}(\sigma^2 \mathbf{I} + r_t^2(\mathbf{x}_t)\mathrm{diag}[\hat{\mathbf{k}}\odot \bar{\hat{\mathbf{k}}}])\mathbf{F} \mathbf{D}_{\downarrow s}^T)^{-1}(\mathbf{y}-\mathbf{A}D_t(\mathbf{x}_t)) \\
        &= \mathbf{F}^{-1}\mathrm{diag}[\bar{\hat{\mathbf{k}}}]\mathbf{F} \mathbf{D}_{\downarrow s}^T(\mathbf{D}_{\downarrow s}\mathbf{F}^{-1}\mathrm{diag}[\sigma^2 + r_t^2(\mathbf{x}_t)\hat{\mathbf{k}}\odot \bar{\hat{\mathbf{k}}}]\mathbf{F} \mathbf{D}_{\downarrow s}^T)^{-1}(\mathbf{y}-\mathbf{A}D_t(\mathbf{x}_t)) 
    \end{align}
    By Lemma~\ref{lemma:FDF}, we have $\mathbf{D}_{\downarrow s}\mathbf{F}^{-1}=\mathbf{F}^{-1}_{\downarrow s}\mathbf{D}_{\Downarrow s}$. Taking the hermitian transpose to both side and leveraging $\mathbf{F}^{-1}=\frac{1}{d}\mathbf{F}^H$ and $\mathbf{F}_{\downarrow s}^{-1}=\frac{1}{d/s}\mathbf{F}_{\downarrow s}^H$, we also have $\mathbf{F} \mathbf{D}_{\downarrow s}^T=s \mathbf{D}_{\Downarrow s}^T \mathbf{F}_{\downarrow s}$. So
    \begin{align}
        \mathbf{v} &= \mathbf{F}^{-1}\mathrm{diag}[\bar{\hat{\mathbf{k}}}]s \mathbf{D}_{\Downarrow s}^T \mathbf{F}_{\downarrow s}(\mathbf{F}^{-1}_{\downarrow s}\mathbf{D}_{\Downarrow s}\mathrm{diag}[\sigma^2 + r_t^2(\mathbf{x}_t)\hat{\mathbf{k}}\odot \bar{\hat{\mathbf{k}}}] s \mathbf{D}_{\Downarrow s}^T \mathbf{F}_{\downarrow s})^{-1}(\mathbf{y}-\mathbf{A}D_t(\mathbf{x}_t)) \\
        &= \mathbf{F}^{-1}\mathrm{diag}[\bar{\hat{\mathbf{k}}}]s \mathbf{D}_{\Downarrow s}^T \mathbf{F}_{\downarrow s}(\mathbf{F}^{-1}_{\downarrow s}\mathrm{diag}[\sigma^2 + r_t^2(\mathbf{x}_t)(\hat{\mathbf{k}}\odot \bar{\hat{\mathbf{k}}})_{\Downarrow s}]  \mathbf{F}_{\downarrow s})^{-1}(\mathbf{y}-\mathbf{A}D_t(\mathbf{x}_t)) \\
        &= \mathbf{F}^{-1}\mathrm{diag}[\bar{\hat{\mathbf{k}}}]s \mathbf{D}_{\Downarrow s}^T \mathbf{F}_{\downarrow s}\mathbf{F}^{-1}_{\downarrow s}\mathrm{diag}[\sigma^2 + r_t^2(\mathbf{x}_t)(\hat{\mathbf{k}}\odot \bar{\hat{\mathbf{k}}})_{\Downarrow s}]^{-1}  \mathbf{F}_{\downarrow s}(\mathbf{y}-\mathbf{A}D_t(\mathbf{x}_t)) \\
        &= \mathbf{F}^{-1}(\bar{\hat{\mathbf{k}}}\odot s\mathbf{D}_{\Downarrow s}^T \frac{\mathbf{F}_{\downarrow s}(\mathbf{y}-\mathbf{A}D_t(\mathbf{x}_t))}{\sigma^2 + r_t^2(\mathbf{x}_t)(\hat{\mathbf{k}}\odot \bar{\hat{\mathbf{k}}})_{\Downarrow s}}) \\
        &= \mathbf{F}^{-1}(\bar{\hat{\mathbf{k}}} \odot_s \frac{\mathbf{F}_{\downarrow_s} (\mathbf{y} - \mathbf{A} D_t(\mathbf{x}_t))}{\sigma^2 + r_t^2(\mathbf{x}_t)( \bar{\hat{\mathbf{k}}} \odot \hat{\mathbf{k}})_{\Downarrow_s}})
    \end{align}
\end{proof}

\subsection{Using Conjugate Gradient Method for General Posterior Covariance}
\label{sec:cg}
For general posterior covariance, the closed-form solution for $\mathbf{v}$ is usually unavailable. However, we can still compute $\mathbf{v}$ efficiently by using numerical solutions of linear equations. Specifically, we first represent $\mathbf{v}$ as
\begin{equation}
    \mathbf{v} = \mathbf{A}^T \mathbf{u}, \ \ \mathbf{u} = (\sigma^2\mathbf{I} + \mathbf{A} \Sigma_t(\mathbf{x}_t) \mathbf{A}^T)^{-1}(\mathbf{y}- \mathbf{A} D_t(\mathbf{x}_t)).
\end{equation}
Note that $\sigma^2\mathbf{I} + \mathbf{A} \Sigma_t(\mathbf{x}_t) \mathbf{A}^T$ is symmetric and positive-definite, and $\mathbf{u}$ can be represented as the solution of the following linear equations:
\begin{equation}
    (\sigma^2\mathbf{I} + \mathbf{A} \Sigma_t(\mathbf{x}_t) \mathbf{A}^T)\mathbf{u} = \mathbf{y}- \mathbf{A} D_t(\mathbf{x}_t).
\end{equation}
Therefore, $\mathbf{u}$ can be computed with acceptable precision using a sufficient number of CG iterates. In the experiments, we use the black box CG method implemented in $\texttt{scipy.sparse.linalg.cg}$\footnote{\url{https://docs.scipy.org/doc/scipy/reference/generated/scipy.sparse.linalg.cg.html}} with \texttt{tol=1e-4} for Convert~(Section~\ref{sec:convert}), DWT-Var~(Section~\ref{sec:ot}) and TMPD covariance. The other corvariances are isotropic, and therefore we use the closed-form solutions derived in Section~\ref{sec:solu-g}.

\section{Additional Experimental Details and Results}
\label{sec:add-exp}

\subsection{Additional Quantitative Results}
\begin{figure*}[!t]
\renewcommand{\baselinestretch}{1.0}
\centering
\includegraphics[width=0.95\textwidth]{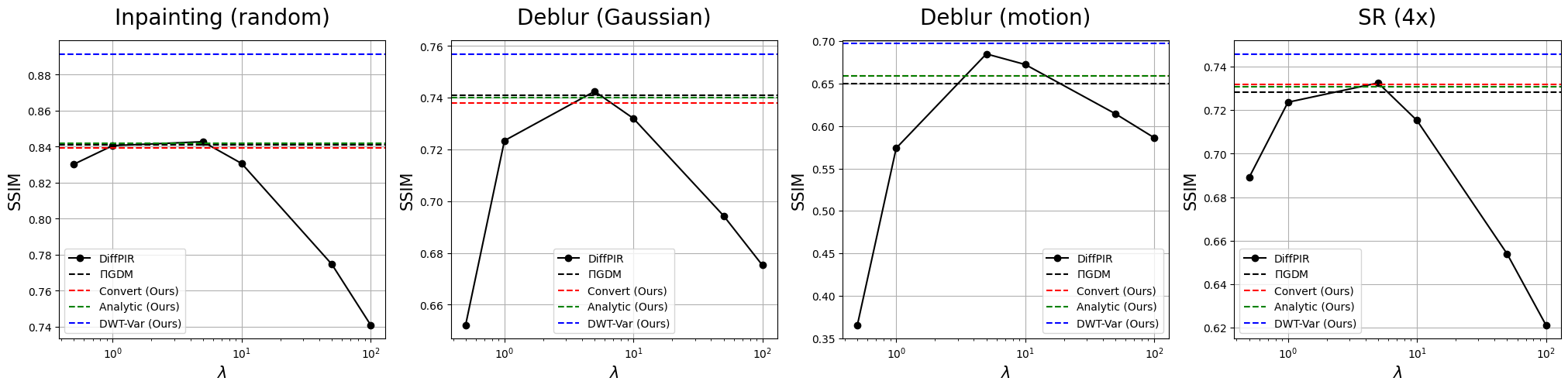}
\caption{\textbf{SSIM comparisons on FFHQ for Type II guidance.} For DiffPIR, we report SSIM under different $\lambda$.}\label{fig:typeII-ssim}
\end{figure*}

\begin{figure*}[!t]
\renewcommand{\baselinestretch}{1.0}
\centering
\includegraphics[width=0.95\textwidth]{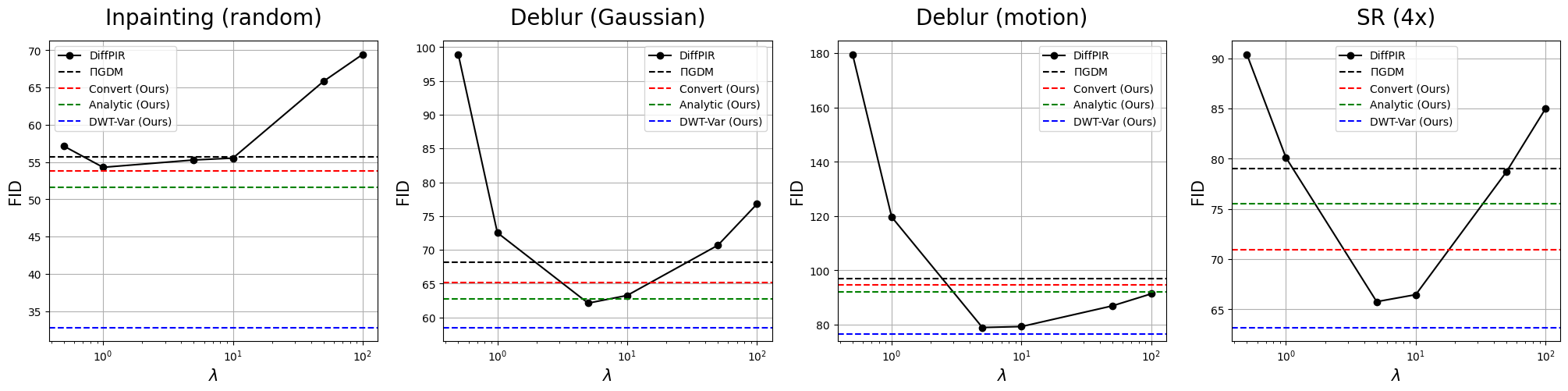}
\caption{\textbf{FID comparisons on FFHQ for Type II guidance.} For DiffPIR, we report FID under different $\lambda$.}\label{fig:typeII-fid}
\end{figure*}

In this section, we report the SSIM~(Figure~\ref{fig:typeII-ssim}) and FID~(Figure~\ref{fig:typeII-fid}) performance for supplementary results of Type II guidance.

\subsection{Training Objective for Latent Variance} \label{app:ot}

In general, the sub-objective of Eq.~\eqref{eq:klobj} at time step $t$ is equivalent to the following MLE objective:
\begin{equation}\label{eq:general-mle-obj}
    \min_{D_t, \Sigma_t} \mathbb{E}_{p_t(\mathbf{x}_0, \mathbf{x}_t)} \left[ (\mathbf{x}_0 - D_t(\mathbf{x}_t) )^T \Sigma_t (\mathbf{x}_t)^{-1} (\mathbf{x}_0 - D_t(\mathbf{x}_t) ) + \log\det \Sigma_t (\mathbf{x}_t) \right]
\end{equation}

For general covariance $\Sigma_t$, evaluating Eq.~\eqref{eq:general-mle-obj} is hard due to the present of the inverse and log-determinant. Nevertheless, when parameterize $\Sigma_t$ using fix orthonormal basis $\mathbf{\Psi}$ as $\Sigma_t(\mathbf{x}_t) = \mathbf{\Psi} \mathrm{diag}[\mathbf{r}_t^2(\mathbf{x}_t)]\mathbf{\Psi}^T$, the inverse and log-determinant possess efficient forms as
\begin{align}
    &\Sigma_t(\mathbf{x}_t)^{-1} = \mathbf{\Psi} \mathrm{diag}[\mathbf{r}_t^{-2}(\mathbf{x}_t)]\mathbf{\Psi}^T, \\
    &\log\det \Sigma_t (\mathbf{x}_t) = \sum_{i=1}^d \log\mathbf{r}_t^2(\mathbf{x}_t)_i.
\end{align}
And the training objective of Eq.~\eqref{eq:general-mle-obj} becomes mirrored to learning diagonal Gaussian posterior (Eq.~\eqref{eq:diag-obj}), but in the transform domain:
\begin{equation}
    \mathbb{E}_{p_t(\mathbf{x}_0, \mathbf{x}_t)}\left[\sum_{i=1}^d \frac{1}{\mathbf{r}_t^2(\mathbf{x}_t)_i}(\mathbf{\Psi}^T\mathbf{x}_0 - \mathbf{\Psi}^TD_t(\mathbf{x}_t))_i^2 + \log \mathbf{r}_t^2(\mathbf{x}_t)_i\right]
\end{equation}

\begin{figure}[!t]
    \centering
    \includegraphics[width=0.8\textwidth]{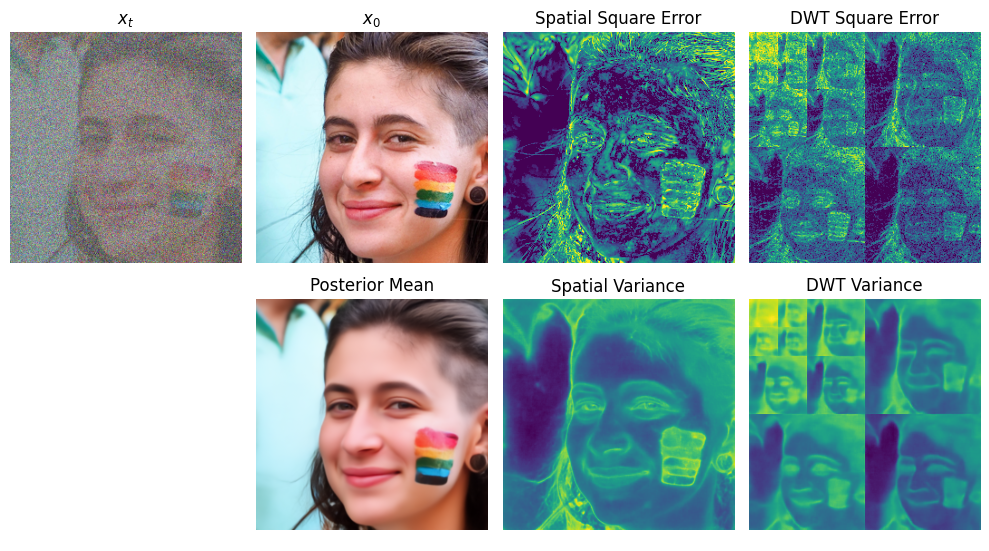}
    \caption{\textbf{Ground truths and model predictions.} $\sigma_t$ is set to 1; Spatial Square Error refers to $(\mathbf{x}_0 - D_t(\mathbf{x}_t))^2$, and DWT Square Error refers to $(\mathbf{\Psi}^T\mathbf{x}_0 - \mathbf{\Psi}^TD_t(\mathbf{x}_t))^2$, where $\mathbf{\Psi}$ is the DWT basis; As can be seen, Spatial Square Error and DWT Square Error can be predicted by Spatial Variance and DWT Variance, respectively.}
    \label{fig:model_ouput}
\end{figure}

We modify unconditional diffusion models of FFHQ dataset from \cite{chung2023diffusion} with nine output channels for predicting three values: 1) posterior mean $\mathbb{E}[\mathbf{x}_0|\mathbf{x}_t]$, 2) $\mathbf{r}^2_t(\mathbf{x}_t)$ when $\mathbf{\Psi}$ is set to identity matrix, referred to as spatial variance, and 3) $\mathbf{r}^2_t(\mathbf{x}_t)$ when $\mathbf{\Psi}$ is set to DWT basis, referred to as DWT variance. Example model predictions see Figure~\ref{fig:model_ouput}. As can be seen, similar to spatial variance, the DWT variance can act as a reliable predictor for the square errors of the DWT coefficients. In fact, it can be shown that the DWT variance is the MMSE estimator of the square errors of DWT coefficients.

\subsection{DDNM as Noiseless DiffPIR}
\label{sec:app-ddnm}
\begin{table}[!t]
\renewcommand{\baselinestretch}{1.0}
\renewcommand{\arraystretch}{1.0}
\centering
    \begin{tabular}{@{}lc@{}}
        \toprule
        \multicolumn{1}{c}{Task} & Avg. MAD \\
        \midrule
        Inpaint (Random) & $1.6975\times 10^{-6}$ \\
        Deblur (Gaussian) & $6.4629 \times 10^{-4}$ \\
        Deblur (Motion) & $1.5461 \times 10^{-3}$ \\
        Super resolution ($4\times$) & $1.1937\times 10^{-2}$ \\
        \bottomrule
    \end{tabular}
    \caption{\textbf{DDNM v.s. DiffPIR in noiseless inverse problems.} We report the averaged MAD between their $\mathbf{x}_0^{(t)}$ averaged over all sampling steps and test images.}
    \label{tab:ddnm-vs-diffpir}
\end{table}
To validate Proposition~\ref{prop:ddnm}, we re-implement DDNM under DiffPIR codebase. DiffPIR deals with noiseless inverse problems by setting $\sigma$ to a relatively low value for $\rho_t$ in Eq.~\eqref{eq:exy-diffpir}~(\emph{e.g.}, 0.001 in DiffPIR codebase). We demonstrate that directly using the DDNM solution in~Eq.~\eqref{eq:ddnm} produces similar results in the noiseless case. Table~\ref{tab:ddnm-vs-diffpir} reports the Mean Absolute Difference~(MAD)\footnote{MAD between $\mathbf{x}$ and $\mathbf{y}$ is defined by $\lVert \mathbf{x} - \mathbf{y} \rVert_1 / d$.} between the conditional posterior means in equations~\ref{eq:exy-diffpir} and \ref{eq:ddnm} by averaging over all the sampling steps and test images. Overall, the MAD is negligible in comparison to the data range of $[-1, 1]$. Debluring and super resolution yield larger MAD than inpainting, since several approximations are made for computing $\mathbf{A}^{\dagger}$, while for inpainting $\mathbf{A}^{\dagger}$ is exact. Details are given as follows:.

\textbf{Inpainting.} We construct the DDNM solution $\mathbb{E}_q[\mathbf{x}_0|\mathbf{x}_t, \mathbf{y}]$ for the inpainting case $\mathbf{A}=\mathbf{D}_\mathbf{m}$ as follows:
\begin{equation}
    \mathbb{E}_q[\mathbf{x}_0|\mathbf{x}_t, \mathbf{y}] = \Tilde{\mathbf{y}} + (1 - \mathbf{m}) \odot D_t(\mathbf{x}_t)
\end{equation}
\begin{proof}
    Since $\mathbf{D}_\mathbf{m} \mathbf{D}_\mathbf{m}^T = \mathbf{I}$ is non-singular, we can directly obtain $\mathbf{A}^\dagger$ by $\mathbf{A}^\dagger=\mathbf{D}_{\mathbf{m}}^T(\mathbf{D}_\mathbf{m} \mathbf{D}_\mathbf{m}^T)^{-1}=\mathbf{D}_{\mathbf{m}}^T$. Plug in Eq.~\eqref{eq:ddnm}, we have
    \begin{align}
        \mathbb{E}_q[\mathbf{x}_0|\mathbf{x}_t, \mathbf{y}] &= \mathbf{D}_{\mathbf{m}}^T \mathbf{y} + (\mathbf{I} - \mathbf{D}_{\mathbf{m}}^T\mathbf{D}_{\mathbf{m}})D_t(\mathbf{x}_t) \\
        &= \Tilde{\mathbf{y}} + (\mathbf{I} - \mathrm{diag}[\mathbf{m}])D_t(\mathbf{x}_t) \\
        &= \Tilde{\mathbf{y}} + \mathrm{diag}[1-\mathbf{m}]D_t(\mathbf{x}_t) \\
        &= \Tilde{\mathbf{y}} + (1-\mathbf{m})\odot D_t(\mathbf{x}_t)
    \end{align}
\end{proof}

\textbf{Debluring.} We construct the pseudo inverse $\mathbf{A}^{\dagger}$ for the linear operator in the debluring case $\mathbf{A}=\mathbf{F}^{-1}\mathrm{diag}[\hat{\mathbf{k}}]\mathbf{F}$ as follows
\begin{equation}
    \mathbf{A}^{\dagger} = \mathbf{F}^{-1}\mathrm{diag}[\hat{\mathbf{k}}]^{\dagger}\mathbf{F}
\end{equation}
where $\mathrm{diag}[\hat{\mathbf{k}}]^{\dagger}$ is defined as $\mathrm{diag}[\hat{\mathbf{k}}]^{\dagger} = \mathrm{diag}[[l_1, l_2, ...]^T]$ with $l_i = 0$ if $\hat{\mathbf{k}}_i = 0$\footnote{Numerically, $|\hat{\mathbf{k}}_i|$ is always larger than zero. We threshold $\hat{\mathbf{k}}_i$ to zero when $|\hat{\mathbf{k}}_i|<3\times 10^{-2}$ similar to \cite{wang2023zeroshot}} otherwise $l_i = 1/\hat{\mathbf{k}}_i$.
\begin{proof}
    It is easy to see that with the above construction, $\mathbf{A}^{\dagger}$ satisfies $\mathbf{A}\mathbf{A}^{\dagger}\mathbf{A} = \mathbf{A}$.
\end{proof}

\textbf{Super resolution.} We directly use \texttt{torch.nn.functional.interpolate} in place of $\mathbf{A}^{\dagger}$ for the super resolution case.

\subsection{Converting Optimal Solutions Between Different Perturbation Kernels}
\label{app:convert kernels}
Suppose we are given a family of optimal solutions $q_t$ defined by the following MLE objectives for all $t\in [0, T]$:
\begin{equation}
\label{eq:q-mle}
    \max_{q_t} \mathbb{E}_{\mathbf{x}_0\sim p(\mathbf{x}_0), \epsilon\sim \mathcal{N}(\mathbf{0}, \mathbf{I})} \log q_t(\mathbf{x}_0|\mathbf{x}_t=s_t(\mathbf{x}_0+\sigma_t \epsilon))
\end{equation}
As can be seen, $q_t$ equals the optimal solutions to Eq.~\eqref{eq:klobj} when the perturbation kernels $p_t(\mathbf{x}_t|\mathbf{x}_0)$ are set to $\mathcal{N}(\mathbf{x}_t|s_t\mathbf{x}_0, s_t^2 \sigma_t^2\mathbf{I})$. Now, suppose we want to perform sampling based on the diffusion ODE~(SDE) under the perturbation kernels $\mathcal{N}(\mathbf{x}_t|\Tilde{s}_t\mathbf{x}_0, \Tilde{s}_t^2 \Tilde{\sigma}_t^2\mathbf{I})$. This means that we are required to provide optimal solutions $\Tilde{q}_t$ defined by the following objectives for all $t\in [0, T]$:
\begin{equation}
\label{eq:tq-mle}
    \max_{\Tilde{q}_t} \mathbb{E}_{\mathbf{x}_0\sim p(\mathbf{x}_0), \epsilon\sim \mathcal{N}(\mathbf{0}, \mathbf{I})} \log \Tilde{q}_t(\mathbf{x}_0|\mathbf{x}_t=\Tilde{s}_t(\mathbf{x}_0+\Tilde{\sigma}_t \epsilon))
\end{equation}

The idea is that, $\Tilde{q}_t$ can be directly represented by $q_t$, so we do not require to perform re-training:
\begin{equation}
\label{eq:q2tq}
    \Tilde{q}_t(\mathbf{x}_0|\mathbf{x}_t=\mathbf{x}) = q_{t'}(\mathbf{x}_0|\mathbf{x}_{t'}=\frac{s_{t'}}{\Tilde{s}_t}\mathbf{x}), \ \ \sigma_{t'} = \Tilde{\sigma}_t
\end{equation}

\begin{proof}
    We only need to show that $\Tilde{q}_t$ defined by Eq.~\eqref{eq:q-mle} and Eq.~\eqref{eq:q2tq} are equivalent to $\Tilde{q}_t$ defined by Eq.~\eqref{eq:tq-mle}. From Eq.~\eqref{eq:q2tq}, we actually know that $q_{t'}(\mathbf{x}_0|\mathbf{x}_{t'}=\mathbf{x})=\Tilde{q}_t(\mathbf{x}_0|\mathbf{x}_t=\frac{\Tilde{s}_t}{s_{t'}}\mathbf{x})$, plug it in Eq.~\eqref{eq:q-mle} at $t'$ we have
    \begin{equation}
        \max_{\Tilde{q}_t} \mathbb{E}_{\mathbf{x}_0\sim p(\mathbf{x}_0), \epsilon\sim \mathcal{N}(\mathbf{0}, \mathbf{I})} \log \Tilde{q}_t(\mathbf{x}_0|\mathbf{x}_t=\Tilde{s}_t(\mathbf{x}_0+\sigma_{t'} \epsilon)), \ \ \sigma_{t'} = \Tilde{\sigma}_t
    \end{equation}
    which is equivalent to Eq.~\eqref{eq:tq-mle}.
\end{proof}

For example, suppose we are given optimal solutions $q_t$ under DDPM perturbation kernels, i.e., $p_t(\mathbf{x}_t|\mathbf{x}_0)=\mathcal{N}(\mathbf{x}_t|\sqrt{\bar{\alpha}_t}\mathbf{x}_0, \bar{\beta}_t\mathbf{I})$.  We aim to convert these solutions to optimal solutions $\Tilde{q}_t$ under perturbation kernels used in Section~\ref{sec:intro dpm}, i.e., $p_t(\mathbf{x}_t|\mathbf{x}_0)=\mathcal{N}(\mathbf{x}_t|\mathbf{x}_0, t^2\mathbf{I})$. We can realize it using following two steps: (1) finding $t'$ such that $\sqrt{\frac{\bar{\beta}_{t'}}{\bar{\alpha}_{t'}}} = t$, and (2) scaling the input of $q_{t'}$ by the factor of $\sqrt{\bar{\alpha}_{t'}}$. Formally,
\begin{equation}
    \Tilde{q}_t(\mathbf{x}_0|\mathbf{x}_t=\mathbf{x}) = q_{t'}(\mathbf{x}_0|\mathbf{x}_{t'}=\sqrt{\bar{\alpha}_{t'}}\mathbf{x}), \ \ \sqrt{\frac{\bar{\beta}_{t'}}{\bar{\alpha}_{t'}}} = t
\end{equation}

\subsection{Qualitative Results}
In this section, we present additional visual examples for qualitative comparisons.




\newpage



\begin{figure}[!t]
    \centering
    \includegraphics[width=0.8\linewidth]{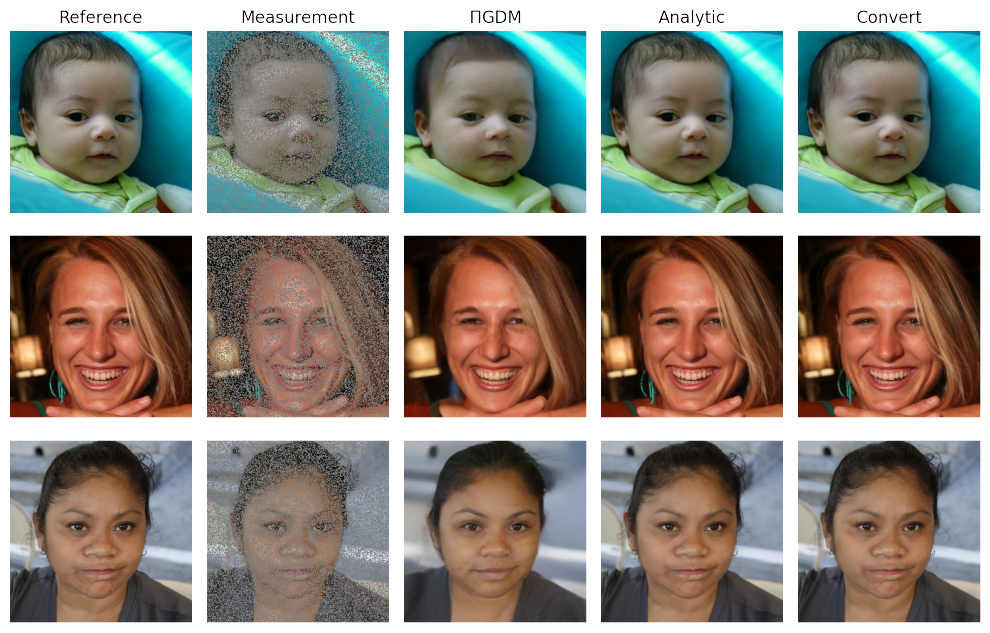}
    \caption{\textbf{Qualitative results for Table~\ref{tab:complete-pgdm} on random inpainting.}}
    \label{fig:complete_pgdm_inpaint}
\end{figure}

\begin{figure}[!t]
    \centering
    \includegraphics[width=0.8\linewidth]{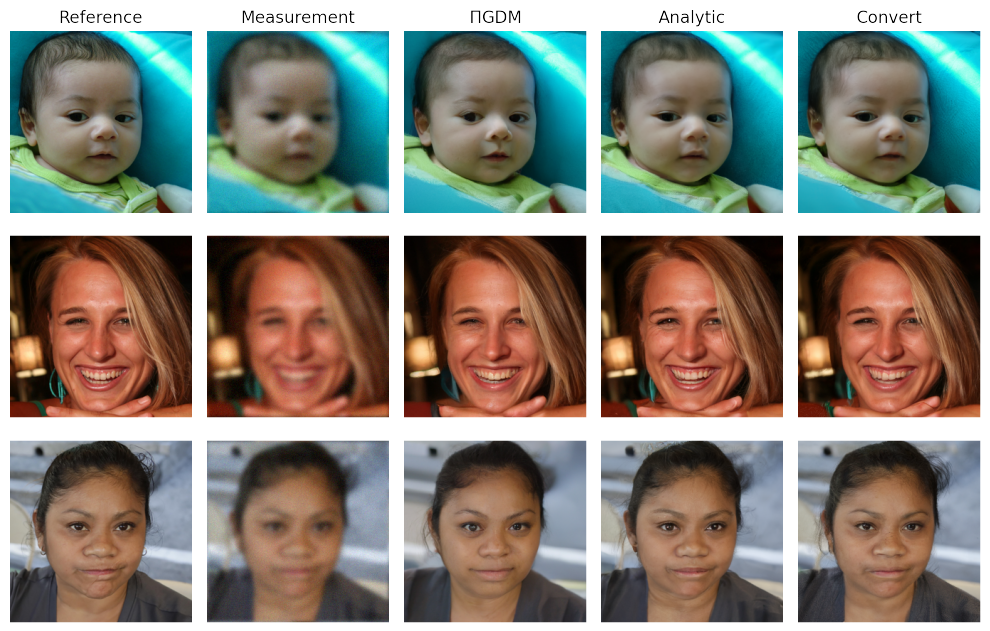}
    \caption{\textbf{Qualitative results for Table~\ref{tab:complete-pgdm} on Gaussian debluring.}}
    \label{fig:complete_pgdm_gaussian_deblur}
\end{figure}

\begin{figure}[!t]
    \centering
    \includegraphics[width=0.8\linewidth]{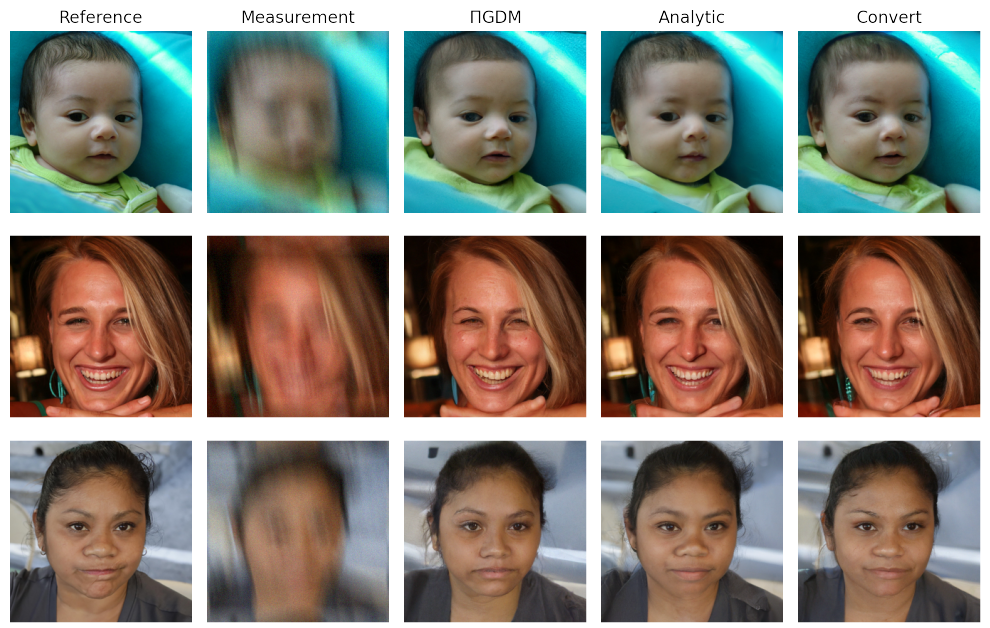}
    \caption{\textbf{Qualitative results for Table~\ref{tab:complete-pgdm} on motion debluring.}}
    \label{fig:complete_pgdm_motion_deblur}
\end{figure}

\begin{figure}[!t]
    \centering
    \includegraphics[width=0.8\linewidth]{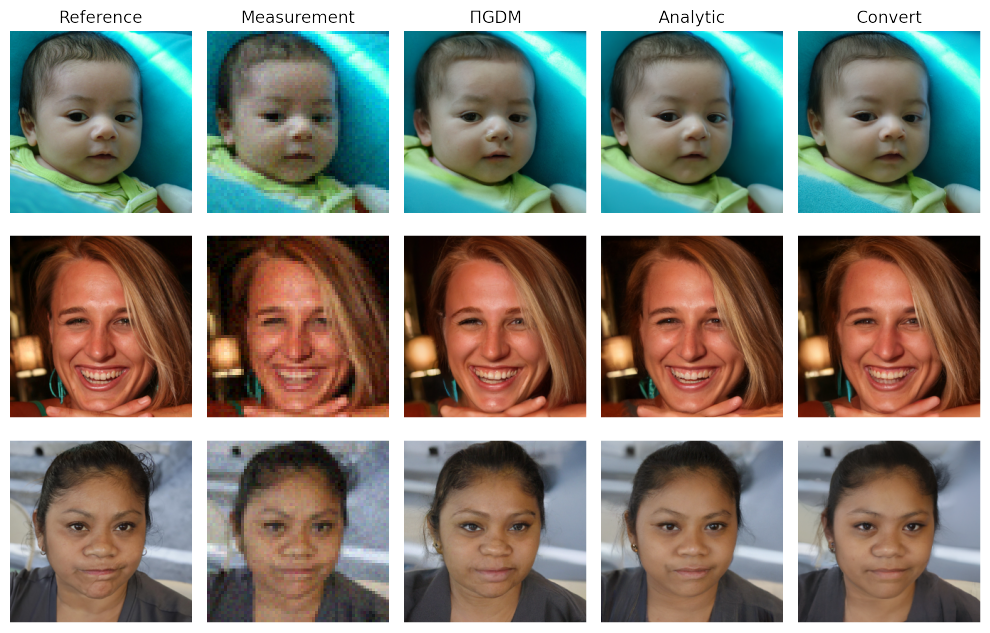}
    \caption{\textbf{Qualitative results for Table~\ref{tab:complete-pgdm} on super resolution.}}
    \label{fig:complete_pgdm_super_resolution}
\end{figure}






\end{document}